\documentclass[letterpaper]{article} 
\usepackage{aaai2026}  
\usepackage{times}  
\usepackage{helvet}  
\usepackage{courier}  
\usepackage[hyphens]{url}  
\usepackage{graphicx} 
\usepackage{natbib}  
\usepackage{caption} 
\usepackage{multirow}
\usepackage{graphicx}
\usepackage{xcolor}         
\usepackage{colortbl} 
\usepackage{booktabs}
\usepackage{amssymb}
\usepackage{amsmath}
\usepackage{cleveref}
\usepackage{subcaption}

\title{Machine Learning for Sustainable Rice Production: Region-Scale Monitoring of Water-Saving Practices in Punjab, India}

\author{
    Ando Shah\textsuperscript{\rm 1}\thanks{Work done while at Microsoft AI for Good Lab, Kenya.}\thanks{Corresponding author}, 
    Rajveer Singh\textsuperscript{\rm 2}, 
    Akram Zaytar\textsuperscript{\rm 3}, 
    Girmaw Abebe Tadesse\textsuperscript{\rm 3},\\ 
    Caleb Robinson\textsuperscript{\rm 3}, 
    Negar Tafti\textsuperscript{\rm 2}, 
    Stephen A. Wood\textsuperscript{\rm 2,4}, 
    Rahul Dodhia\textsuperscript{\rm 3}, 
    Juan M. Lavista Ferres\textsuperscript{\rm 3}\\
}

\affiliations {
    \textsuperscript{\rm 1}University of California, Berkeley; \textsuperscript{\rm 2}The Nature Conservancy\\
\textsuperscript{\rm 3}Microsoft AI for Good Research Lab; \textsuperscript{\rm 4}Yale School of the Environment \\
  ando@berkeley.edu
}

\begin{document}

\maketitle

\begin{abstract}
Rice cultivation supplies half the world's population with staple food, while also being a major driver of freshwater depletion--consuming roughly a quarter of global freshwater--and accounting for $\sim$48\% of greenhouse gas emissions from croplands. In regions like Punjab, India, where groundwater levels are plummeting at 41.6 cm/year, adopting water-saving rice farming practices is critical. Direct-Seeded Rice (DSR) and Alternate Wetting and Drying (AWD) can cut irrigation water use by 20–40\% without hurting yields, yet lack of spatial data on adoption impedes effective adaptation policy and climate action. We present a machine learning framework to bridge this data gap by monitoring sustainable rice farming at scale. In collaboration with agronomy experts and a large-scale farmer training program, we obtained ground-truth data from $\sim$1,400 fields across Punjab. Leveraging this partnership, we developed a novel dimensional classification approach that decouples sowing and irrigation practices, achieving F1 scores of 0.8 and 0.74 respectively, solely employing Sentinel-1 satellite imagery. Explainability analysis reveals that DSR classification is robust while AWD classification depends primarily on planting schedule differences, as Sentinel-1's 12-day revisit frequency cannot capture the higher frequency irrigation cycles characteristic of AWD practices. Applying this model across 3 million fields reveals spatial heterogeneity in adoption at the state level, highlighting gaps and opportunities for policy targeting. Our district-level adoption rates correlate well with government estimates (Spearman's $\rho$=0.69 and Rank Biased Overlap=0.77). This study provides policymakers and sustainability programs a powerful tool to track practice adoption, inform targeted interventions, and drive data-driven policies for water conservation and climate mitigation at regional scale.
\end{abstract}

\begin{links}
    \link{Code \& Dataset}{https://github.com/microsoft/rice-irrigation-mapping-s1s2}
    \link{Extended version}{https://arxiv.org/abs/2507.08605}
\end{links}

\section{Introduction}\label{sec:introduction}

Rice is a staple food for half of the world's population, but traditional paddy cultivation requires enormous quantities of water—2,000–5,000 liters per kilogram of rice produced~\cite{bouman2007water}, consuming 24-30\% of world's freshwater~\cite{surendran2021use}. As climate change intensifies water scarcity in many regions, sustainable water management in rice cultivation becomes essential for ensuring long-term food security, critical to addressing the United Nations Zero Hunger Sustainable Development Goal (SDG). Furthermore, rice cultivation accounts for $\sim$48\% of greenhouse gas (GHG) emissions from croplands as a direct consequence of water-intensive paddy cultivation~\cite{qian2023greenhouse}. Two water-saving practices have emerged as particularly promising: Direct Seeded Rice (DSR) and Alternate Wetting and Drying (AWD). DSR eliminates the need to raise and transplant seedlings by directly sowing seeds into the field, reducing water consumption by 20-40\%~\cite{bhushan2007saving,bouman2007water}, relative to traditionally practiced puddled transplanted rice (PTR). AWD involves periodically allowing fields to dry before re-flooding, reducing water use by up to 30\% compared to traditional continuous flooding (CF)~\cite{richards2014alternate}. When properly managed, these techniques maintain or even enhance yields, presenting a balanced approach for sustainable rice production~\cite{lansing2023adaptive}.

Despite the promise of these practices, their scale of adoption remains poorly understood. This knowledge gap severely hampers evidence-based policymaking, resource allocation, and intervention planning for several reasons:

\begin{itemize}
\item Ground surveys are labor-intensive, expensive, and often constrained by limited access to remote fields
\item Manual methods struggle to differentiate between traditional and newly adopted practices
\item Adoption patterns are not routinely tracked in national or sub-national surveys
\item Understanding current adoption helps target resources where they're most needed
\end{itemize}

\begin{figure*}[!tbhp]
\centering
\includegraphics[width=0.82\linewidth]{inference-main_v3.2.pdf}
\caption{DSR predictions for Punjab in 2024. A) Map of Punjab with predicted DSR adoption distributions, shown with district boundaries; darker shades denote higher DSR plot-density. B) Comparison between model predictions and Punjab government estimates of DSR adoption by district, ordered by acres of DSR activity. C) District-level adoption compared to government records; evaluated using Spearman’s $\rho$=0.69, R$^2$=0.46, Rank Biased Overlap=0.77 and MAE$ \approx12$ thousand acres. Key districts are labeled.}
\label{fig:infer2}
\end{figure*}

Remote sensing offers a potential solution for monitoring water management practices at scale. While satellite-based methods have successfully mapped rice cultivation using multispectral optical and synthetic aperture radar (SAR) approaches~\cite{nguyen2016mapping, fatchurrachmanHighResolutionMappingPaddy2022, singhaHighResolutionPaddy2019}, detecting specific water management practices remains challenging. Existing studies rely heavily on prior knowledge of planting dates~\cite{fikriyahDiscriminatingTransplantedDirect2019, villano2019separability} or use coarse resolution data~\cite{gumma2015mapping}, limiting their scalability in regions where planting dates vary significantly, or smallholder-dominated regions.

We focused our analysis on Punjab, India, a major rice-producing region facing severe water stress. Punjab contributes 10.42\% of India's rice production~\cite{MARKETINTELLIGENCEREPORT2024}, but faces groundwater depletion of 41.6 cm/year~\cite{baweja2017groundwater}. This crisis stems from the region's semi-arid climate, rice-wheat cropping systems that increased irrigation demand, and groundwater extraction facilitated by free electricity~\cite{gupta2023free, PunjabStateAgriPolicy2023}. Approximately 73\% of irrigation water comes from rapidly depleting groundwater sources~\cite{sidhu2021spatio}.
To address these challenges, we leveraged data from The Nature Conservancy's Promoting Regenerative and No-burn Agriculture (PRANA) project\footnote{\url{https://www.nature.org/en-us/about-us/where-we-work/india/our-priorities/prana/}}. PRANA trained approximately 150,000 farmers across Punjab on sustainable water management methods, while collecting detailed field-level data from about 1,400 participants. This included documentation of sowing methods, irrigation practices, and precise field boundaries--critical ground truth data for our remote sensing approach.
Our study introduces a dimensional framework for monitoring water management practices using Sentinel-1 (S1) satellite imagery. By separating practices along sowing (DSR vs.~PTR) and irrigation (AWD vs.~CF) dimensions, we leverage their distinct temporal signatures while avoiding the limitations of simultaneous detection. Conceptually, this framing clarifies that agronomic processes driving sowing and irrigation decisions are distinct and temporally separable. Importantly, our approach does not require prior knowledge of cropping calendars, enabling large-scale monitoring, with results shown in \Cref{fig:infer2}. Our main contributions include:

\begin{itemize}
    \item \textit{Methodologically}, we develop a comparative evaluation framework that operationalizes this decomposition at state scale, integrating features extracted from S1 using traditional methods and pretrained Earth Observation (EO) models.
    \item \textit{Empirically}, we show that this formulation achieves higher accuracy than a three-class model (0.798 v/s 0.616). We conduct explainability analysis of model outputs to provide evidence that high irrigation classification performance (0.742 F1) stems from planting schedule artifacts rather than irrigation patterns of frequencies higher than can be currently captured by a single S1 satellite.
    \item \textit{Practically} we demonstrate scalable and near-operational monitoring of $>$3 million agricultural plots in Punjab, with aggregated district-level predictions correlating strongly with government records.
\end{itemize}

\section{Background}

\subsection{Study Area}

\begin{figure}[!t]
\centering
\includegraphics[width=\linewidth]{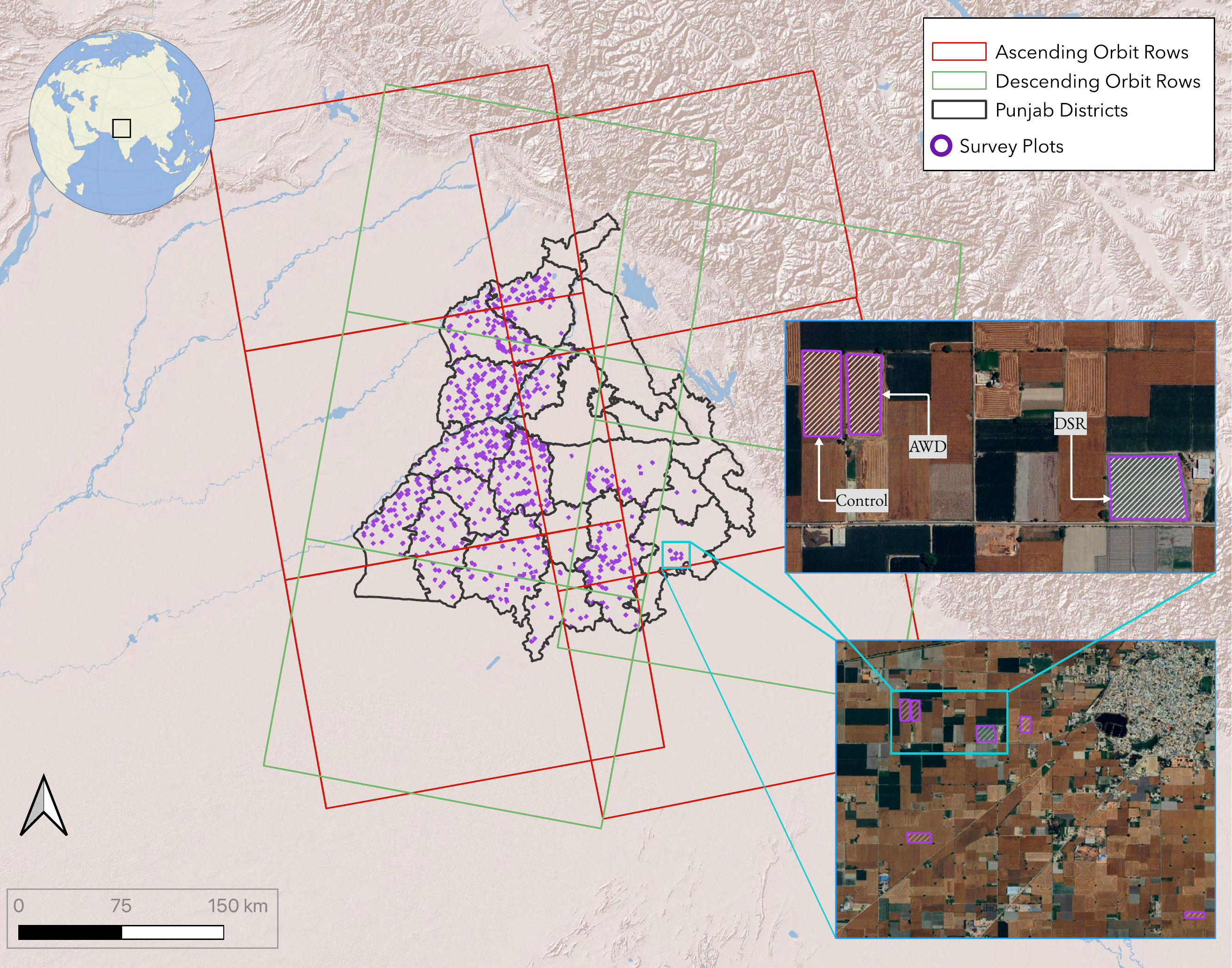}
\caption{Map of Punjab showing the study area, Sentinel-1 satellite coverage (ascending and descending orbits), and locations of surveyed plots. Insets show sample plots at different zoom levels, with the innermost inset displaying delineated fields from three classes: Control (plots that use puddled transplanting and continuous flooding), alternate wetting and drying (AWD), and direct seeded rice (DSR). Background satellite imagery from Bing Maps.}
\label{fig:punjab-state}
\end{figure}

Our study focuses on Punjab, India (Fig. \ref{fig:punjab-state}), a major rice-producing region facing severe water stress. Punjab contributes 10.42\% of India's rice production and approximately 3.3\% of global output~\cite{MARKETINTELLIGENCEREPORT2024}, but faces rapid groundwater depletion. This crisis stems from wheat-rice double-cropping system using PTR and CF methods~\cite{han2022annual} and groundwater extraction facilitated by free electricity, with 73\% of irrigation extracted from rapidly depleting underground aquifers. A 2022 Government report classified 114 of Punjab's 150 blocks as over-exploited~\cite{CGWBreport2022}. To address these challenges, we partnered with The Nature Conservancy's PRANA project, which trained approximately 150,000 farmers on sustainable water management while collecting detailed field-level data from more than 1,400 participants.

\begin{figure*}[t]
    \begin{minipage}[t]{0.48\textwidth}
        \centering
        \includegraphics[width=\linewidth]{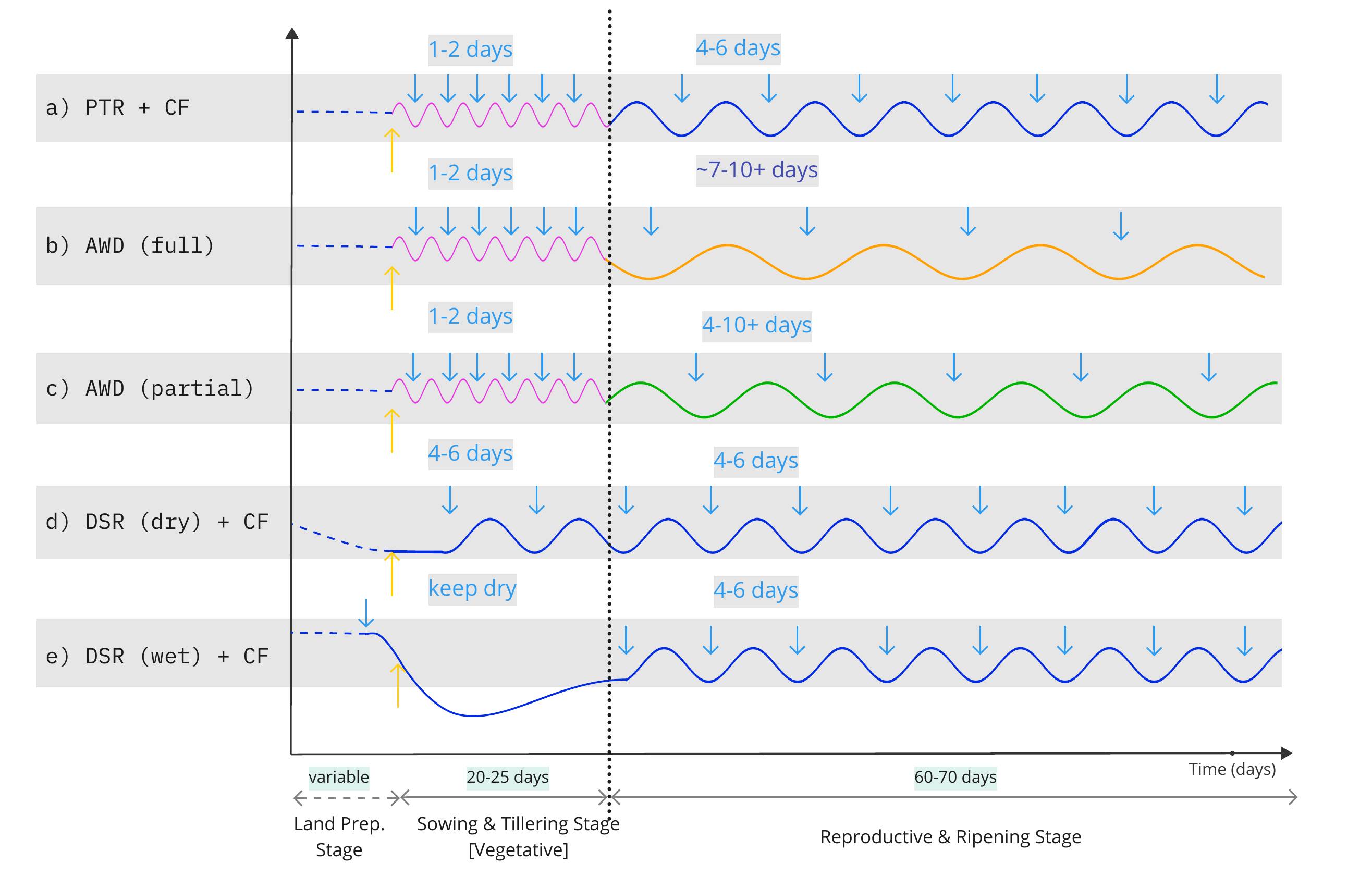}
        \caption{Water management practices across classes, created in collaboration with rice agronomy experts. The $y$-axis represents water level in a rice paddy, and $x$-axis shows growth stages. The growing season naturally separates into two phases: initial (left of dotted line) and main (right), which have distinct irrigation needs. Blue arrows indicate irrigation events, while yellow arrows denote sowing. The different practices create distinct temporal signatures, with the vertical dotted line marking the boundary between sowing and irrigation dimensions.}
        \label{fig:temporal}
    \end{minipage}
    \hfill
    \begin{minipage}[t]{0.48\textwidth}
        \centering
        \includegraphics[width=\linewidth]{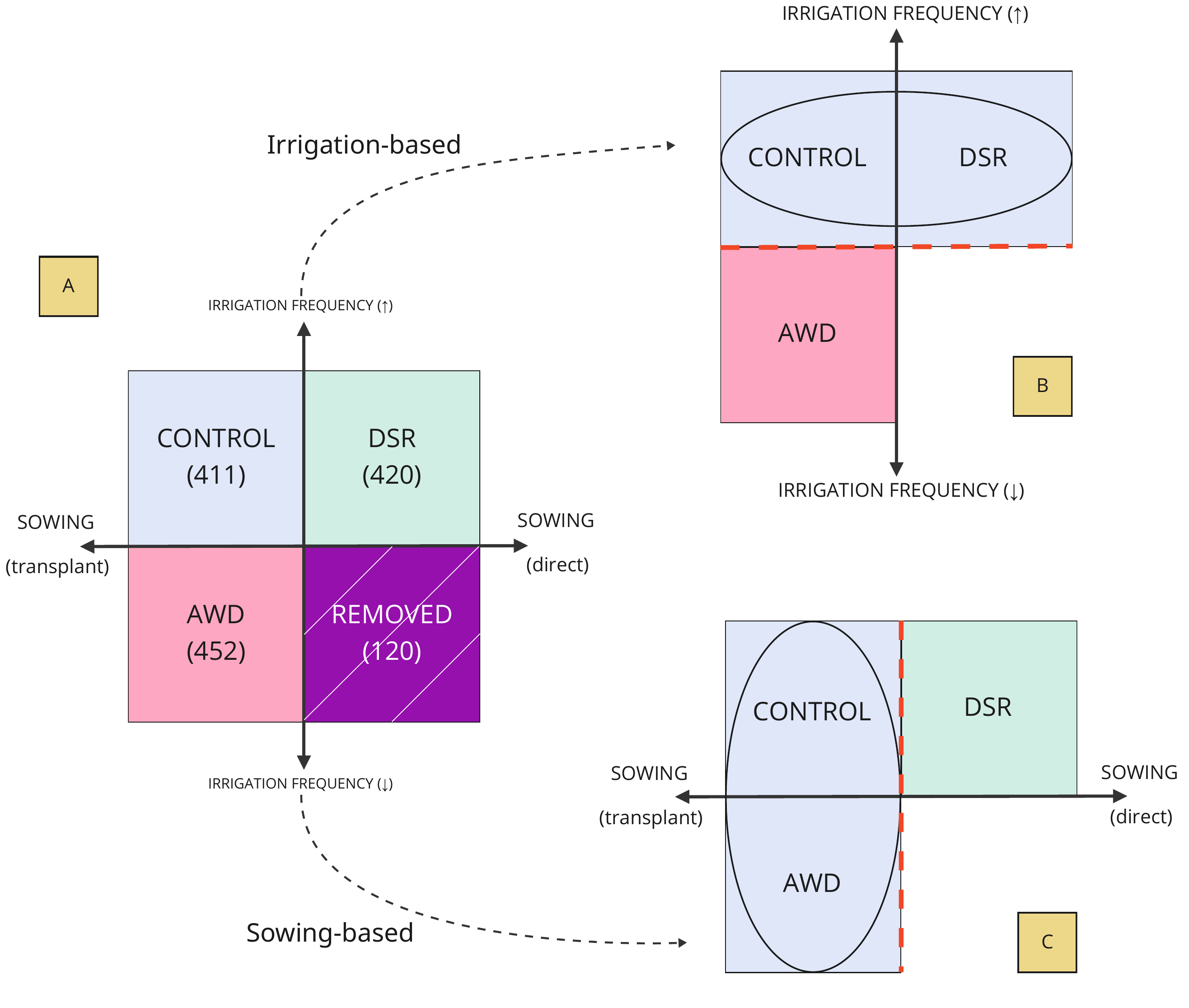}
        \caption{Dimensional divisions of dataset by practice type. A) Plots divided by sowing and irrigation dimensions, with number of plots in parentheses. B) Classification of irrigation practices (grouping Control and DSR as CF). C) Classification of sowing practices (grouping Control and AWD as PTR). Dotted red lines in B) and C) show classes are separated in a task-dependent manner.}
        \label{fig:class-def}
    \end{minipage}
\end{figure*}

\subsection{Challenges}
\label{sec:problem-def}

While satellite-based monitoring offers a potential solution, several technical challenges have limited its application:

\subsubsection{Water Management Practices and Separability:}

Water-saving practices in rice cultivation occur at different crop stages and create distinct water level patterns. DSR affects water management during early sowing and tillering phases, while AWD impacts water dynamics from tillering through harvest. These practices create unique temporal signatures in water levels and soil moisture that can be detected via SAR remote sensing~\cite{ijgi8070312, fikriyahDiscriminatingTransplantedDirect2019, villano2019separability, gumma2015mapping}. As shown in \Cref{fig:temporal}, PTR maintains consistently high water levels, while DSR shows a sharp initial decline after pre-sowing irrigation. Similarly, CF maintains flooded fields throughout the season, whereas AWD creates cyclical wet-dry patterns.

\subsubsection{Dimensional Classification Approach:}

Attempting to classify all water management combinations simultaneously poses substantial challenges due to overlapping characteristics. We address this by separating practices into two independent dimensions: 
\textbf{sowing dimension}, distinguishing between DSR and PTR practices
and \textbf{irrigation dimension}, separating AWD from CF practices.

\subsubsection{Planting Date Heterogeneity:}

Satellite-based monitoring faces significant variation in planting schedules. Previous studies relied on synchronizing satellite data with known planting dates~\cite{fikriyahDiscriminatingTransplantedDirect2019, villano2019separability} – a critical challenge for monitoring millions of plots. In our study area, planting dates span up to 110 days, with DSR plots typically sown first, followed by control plots and AWD plots (Fig. S3). 

\subsubsection{Plot-level Predictions:}
To provide fine-grained insights, we must be able to both identify individual plot boundaries as well as make predictions about these monolithic entities. We employ Fields of The World (FTW) algorithm~\cite{kerner2024fields} to automate the extraction of plot boundaries, enabling us to make predictions at an individual plot level. \\

\section{Materials and Methods}

Let X = $[x_t]_{t=1}^T$ be the seasonal S1 feature sequence for a plot. We want to learn two predictive functions: sowing $f_{s}:X \rightarrow \hat{y}_{s}\in\{\text{DSR,PTR}\}$ and irrigation $f_{i}:X \rightarrow \hat{y}_{i}\in\{\text{AWD,CF}\}$ as parallel binary tasks, additionally fused into a joint prediction, $\hat{y}_{joint} = g(\hat{y}_{s}, \hat{y}_{i})$. We aim to learn these predictors under limited ground truth and heterogeneous planting dates.

\subsection{Ground Truth Data Collection}
We partnered with the PRANA project to obtain reliable training data. Field teams collected detailed plot-level data from intervention sites across 18 districts in Punjab during the 2024 \textit{kharif} rice season. The resulting dataset includes 452 AWD plots, 420 DSR plots and 411 PTR plots employing CF, with mean areas of 4.2, 2.7 and 2.1 acres, respectively (see \Cref{fig:punjab-state} for samples). Sowing and transplantation dates, irrigation schedules, and precise field boundaries were documented through on-site visits. Field polygons were digitized from 2024 Google Earth Pro imagery (30–50 cm) enforcing a minimum area of 2000 m$^2$, and validated in-person for a subset of plots. A negative 10m buffer was applied to mitigate co-registration and boundary errors between the Google Earth basemap and co-registered S1 tiles. 

\subsection{Satellite Data and Processing}
\label{sec:sat-data}

We leveraged S1 SAR data to monitor water management practices across the growing season (April - December 2024). We acquired 108 images from ascending orbit covering Punjab, focusing on May 1 - December 15\footnote{Through experimentation, the ascending orbit dataset was shown to have higher performance and was selected for training and inference}. Previous studies have shown SAR's effectiveness in capturing diverse rice agriculture metrics, including cultural type classification, phenological parameter detection, growing season estimation~\cite{clauss2018estimating, yangSystematicMethodSpatiotemporal2021,sahDiscriminationMonitoringRice2023}, and differentiation between multiple cropping cycles and field-level tillage practices~\cite{singhaHighResolutionPaddy2019, liu2022using}. SAR data was also selected for its cloud-penetrating capabilities during the monsoon season.

\begin{table*}[!ht]
\centering
\resizebox{\textwidth}{!}{
\begin{tabular}{@{}l c c c c c c cc cc cc cc @{}}
\toprule
\multicolumn{6}{c}{\textbf{Features}} & \multicolumn{1}{c}{} & \multicolumn{4}{c}{\textbf{3-Class}} & \multicolumn{4}{c}{\textbf{2-Class}} \\
\cmidrule(r){1-7}\cmidrule(lr){8-11}\cmidrule(l){12-15}
Group & HC & DTW & FT & AE & P & \#Feat. 
& \multicolumn{2}{c}{One-shot} & \multicolumn{2}{c}{Fused}
& \multicolumn{2}{c}{Sowing} & \multicolumn{2}{c}{Irrigation} \\
\cmidrule(lr){8-9}\cmidrule(lr){10-11}\cmidrule(lr){12-13}\cmidrule(lr){14-15}
 &  &  &  &  &  & 
 & Acc & F1 & Acc & F1 & Acc & F1 & Acc & F1 \\
\midrule
Random & $\times$ & $\times$ & $\times$ & $\times$ & $\times$ & 
 & 0.333 & 0.332 & 0.333 & 0.334 & 0.559 & 0.559 & 0.542 & 0.540 \\
\midrule
\multirow{3}{*}{Baselines}
 & $\checkmark$ &  &  &  &  & 61
 & 0.574 & 0.567 & 0.527* & 0.534* & 0.775* & 0.769* & 0.698* & 0.687* \\
 &  & $\checkmark$ &  &  &  & 3
 & 0.450 & 0.450 & 0.434 & 0.431 & 0.636 & 0.639 & 0.612 & 0.612 \\
 &  &  & $\checkmark$ &  &  & 96
 & 0.496 & 0.496 & 0.496 & 0.502 & 0.744 & 0.742 & 0.721* & 0.707* \\
\midrule
\multirow{3}{*}{AE}
 &  &  &  & $\checkmark$ &  & 64
 & 0.450 & 0.450 & 0.450 & 0.453 & 0.721 & 0.715 & 0.635 & 0.625 \\
 & $\checkmark$ &  &  & $\checkmark$ &  & 125
 & \textbf{0.620} & \textbf{0.616} & 0.574 & 0.578 & \textbf{0.798} & \textbf{0.796} & 0.713* & 0.697* \\
 &  &  & $\checkmark$ & $\checkmark$ &  & 160
 & 0.589 & 0.588 & 0.597 & 0.601 & \underline{0.783} & \underline{0.774} & \textbf{0.744} & \textbf{0.742} \\
\midrule
\multirow{3}{*}{Presto}
 &  &  &  &  & $\checkmark$ & 6272
 & 0.535* & 0.526* & 0.581 & 0.583 & 0.760 & 0.755 & 0.721 & 0.724 \\
 & $\checkmark$ &  &  &  & $\checkmark$ & 6333
 & 0.550 & 0.541 & 0.567 & 0.567 & 0.744 & 0.745 & \underline{0.736} & \underline{0.739} \\
 &  &  & $\checkmark$ &  & $\checkmark$ & 6368
 & 0.543* & 0.534 & 0.589 & 0.589 & 0.752 & 0.750 & 0.721 & 0.725 \\
\midrule
\multirow{2}{*}{AE + P}
 &  &  &  & $\checkmark$ & $\checkmark$ & 6336
 & 0.581* & 0.575* & 0.581 & 0.583 & 0.752 & 0.750 & 0.736 & 0.736 \\
 &  &  & $\checkmark$ & $\checkmark$ & $\checkmark$ & 6432
 & 0.581* & 0.576* & \textbf{0.605} & \textbf{0.603} & 0.752 & 0.750 & 0.729 & 0.734 \\
\bottomrule
\end{tabular}
}
\caption{Classification performance by dimension of separation. Dimensional separation consistently outperforms 3-class classification for all tasks. Features: HC = hand crafted, DTW = Dynamic Time Weighting + kNN, FT = Fourier transform, P = Presto, AE = AlphaEarth. Results are from LightGBM unless *=RF. \textbf{Bold} indicates best results per task; \underline{underlined}, second-best. All tasks use the best date interval from ablations, while AE features are fixed for the calendar year 2024. `One-shot' refers to a 3-class classification which will mechanically have lower scores, and hence we compare to `fused' which is the combination of making a DSR prediction first, followed by AWD for a given plot. Acc = accuracy, F1 = F1-weighted. \#Feat. indicates the number of features used for that task.}
\label{tab:results-main}
\end{table*}

All imagery was processed using the SNAP toolbox~\cite{filipponi2019sentinel} to extract time series at 10-meter resolution. We extracted VV (Vertical transmit, Vertical receive) and VH (Vertical transmit, Horizontal receive) polarization bands, applying standard preprocessing: orbit file application, noise removal, radiometric calibration, multi-temporal speckle filtering~\cite{yommy2015sar}, and terrain correction to obtain terrain-geocoded backscatter coefficient, $\gamma_0$. For each plot, we calculated mean backscatter values per timestep, generating plot-level timeseries. To handle temporal gaps, we applied temporal smoothing with a 1-D Gaussian filter followed by third-degree polynomial spline fitting, enabling regular sampling despite variable acquisition dates, crucial for dealing with missing timeseries data. We additionally create a VV/VH band ratio time series, which has proven effective for rice mapping in previous studies~\cite{villano2019separability, bazziMappingPaddyRice2019, singha2019high}.

\subsection{Feature Engineering}

We derive complementary feature families that summarize both explicit temporal structure and learned representations of radar backscatter dynamics throughout the rice-growing season. All features are computed at the plot level from smoothed S1 VV, VH, and VV/VH ratio time series.

\noindent \textbf{Analytical Features:} For each polarization (VV, VH, VV/VH), we extract temporal characteristics that include the position and amplitude of the first $k$ troughs, crests, and inflection points; the total number of troughs and crests; summary statistics (mean, minimum, maximum amplitude); parameters from Gaussian kernel fitting to the VV/VH ratio (amplitude, peak day, standard deviation and $R^2$\cite{bazziMappingPaddyRice2019}); and radar vegetation index (RVI) values~\cite{nasirzadehdizaji2019sensitivity} for a total of 61 features, which we refer to as \textit{hand-crafted} (HC) features. To complement these direct time-domain descriptors, we compute Fourier-domain features using fast Fourier transform (FT) that capture periodic irrigation-drainage patterns and overall signal smoothness.

\noindent \textbf{Learnt Features:} We include pretrained representations from two EO foundation models. The Pretrained Remote Sensing Transformer (Presto)~\cite{tseng2023lightweight} encodes up to 24 Sentinel-1 timesteps into 128-dimensional temporal embeddings that capture non-linear phenological trajectories. The Google AlphaEarth (AE) embeddings product~\cite{brown2025alphaearth} provides lightweight 64-dimensional embeddings summarizing annual multisensor (S1, S2, Landsat) variability, capturing landscape level dynamics at 10-meter resolutions. Both these embeddings are inexpensive to acquire, and therefore very attractive to practitioners.

\subsection{Classification Approach}
We test two ensemble learning methods—Random Forest (RF) and Gradient Boosted Trees (GB)--for their proven effectiveness in agricultural classification tasks~\cite{liuUsingSentinel1Sentinel22022, fangComprehensiveReviewRice2024a}. These models effectively handle the heterogeneous feature sets while remaining computationally efficient. We also include a Dynamic Time Warping (DTW) baseline to benchmark performance without explicit feature extraction. DTW directly compares raw S1 time series using dynamic alignment and classifies them via a 1-nearest-neighbor scheme implemented using the tslearn package~\cite{zhang2017dynamic}.

Following our dimensional framework, we structure three classification tasks: a) AWD vs. DSR vs. Control (null hypothesis: all practices on same dimension), b) DSR vs. PTR (sowing dimension), and c) AWD vs. CF (irrigation dimension). For each task, we split data into non-overlapping train/test sets (90:10 ratio) using stratified sampling. We optimize hyperparameters using Optuna~\cite{akiba2019optuna} and evaluate performance using overall accuracy and F1 scores (globally weighted averages). We test a set of combinations of features and classification methods, summarized in \Cref{tab:results-main}. This approach tests our hypothesis that separating water management practices along their natural dimensions (sowing and irrigation) produces more accurate results than attempting simultaneous multi-class classification (i.e. the null hypothesis, 3-class one-shot). For comparison, we also evaluate a fused three-class setup where the sowing and irrigation outputs are combined into a single categorical label via rule-based fusion ($\text{fused} :=\text{DSR}\rightarrow \text{DSR}$, else irrigation decides AWD vs CF) - illustrating compounded error from sequential binary decisions and serves as a fairer baseline.
All results use stratified random splits within the same region-year (Punjab-2024). Please see 
Figure A3 for the complete training workflow.

\renewcommand{\arraystretch}{1.1}
\begin{table*}[!t]
\centering
\resizebox{\textwidth}{!}{%
\begin{tabular}{|clllllllll|l|ll|cc|ccc|}
\hline
\multicolumn{10}{|c|}{Temporal Ranges} &
  \multicolumn{1}{c|}{} &
  \multicolumn{2}{c|}{Sowing} &
  \multicolumn{2}{c|}{Irrigation} &
  \multicolumn{3}{c|}{F1 (weighted)} \\ \cline{1-10} \cline{12-18} 
\multicolumn{1}{|c|}{Row} &
  \multicolumn{1}{c}{Dates} &
  M &
  J &
  J &
  A &
  S &
  O &
  N &
  D &
  \multicolumn{1}{c|}{\multirow{-2}{*}{Lag (\%)}} &
  \multicolumn{1}{c}{DSR} &
  \multicolumn{1}{c|}{PTR} &
  CF &
  AWD &
  ALL &
  DSR &
  AWD \\ \hline
\multicolumn{1}{|c|}{1} &
  May 1 - Aug 15* &
  \cellcolor[HTML]{34A853} &
  \cellcolor[HTML]{34A853} &
  \cellcolor[HTML]{34A853} &
  \cellcolor[HTML]{FBBC04} &
   &
   &
   &
   &
  $\sim$50\% &
  $\checkmark$  &
  $\checkmark$ &
  $>$ &
  $\times$ &
  0.519 &
  0.740 &
  0.718 \\ \hline
\multicolumn{1}{|c|}{2} &
  Jun 1 - Aug 30 &
   &
  \cellcolor[HTML]{34A853} &
  \cellcolor[HTML]{34A853} &
  \cellcolor[HTML]{34A853} &
  \cellcolor[HTML]{FBBC04} &
   &
   &
   &
  $\sim$50\% &
  $>$ &
  $\checkmark$ &
  $\checkmark$ &
  $<$ &
  0.539 &
  0.720 &
  0.723 \\ \hline
\multicolumn{1}{|c|}{3} &
  Jun 1 - Sep 5* &
   &
  \cellcolor[HTML]{34A853} &
  \cellcolor[HTML]{34A853} &
  \cellcolor[HTML]{34A853} &
  \cellcolor[HTML]{FBBC04} &
   &
   &
   &
  $\sim$50\% &
  $>$ &
  $\checkmark$ &
  $\checkmark$ &
  $<$ &
  0.539 &
  \textbf{0.755} &
  0.723 \\ \hline
\multicolumn{1}{|c|}{4} &
  Jun1 - Oct 15 &
   &
  \cellcolor[HTML]{34A853} &
  \cellcolor[HTML]{34A853} &
  \cellcolor[HTML]{34A853} &
  \cellcolor[HTML]{34A853} &
  \cellcolor[HTML]{FBBC04} &
   &
   &
  $\sim$60\% &
  $>$ &
  $\checkmark$ &
  $\checkmark$ &
  $\checkmark$ &
  \textbf{0.543} &
  \textbf{0.751} &
  0.731 \\ \hline
\multicolumn{1}{|c|}{5} &
  July 1 - Oct 15 &
   &
   &
  \cellcolor[HTML]{34A853} &
  \cellcolor[HTML]{34A853} &
  \cellcolor[HTML]{34A853} &
  \cellcolor[HTML]{FBBC04} &
   &
   &
  $\sim$50\% &
  $\times$ &
  $\checkmark$ &
  $\checkmark$ &
  $\checkmark$ &
  0.531 &
  0.728 &
  0.719 \\ \hline
\multicolumn{1}{|c|}{6} &
  Aug 1 - Oct 15 &
   &
   &
   &
  \cellcolor[HTML]{34A853} &
  \cellcolor[HTML]{34A853} &
  \cellcolor[HTML]{FBBC04} &
   &
   &
  $\sim$33\% &
  $\times$ &
  $\times$ &
  $\checkmark$ &
  $\checkmark$ &
  0.489 &
  0.696 &
  0.709 \\ \hline
\multicolumn{1}{|c|}{7} &
  Aug 1 - Nov 15 &
   &
   &
   &
  \cellcolor[HTML]{34A853} &
  \cellcolor[HTML]{34A853} &
  \cellcolor[HTML]{34A853} &
  \cellcolor[HTML]{FBBC04} &
   &
  $\sim$50\% &
  $\times$ &
  $\times$ &
  $\checkmark$ &
  $\checkmark$ &
  0.434 &
  0.728 &
  0.707 \\ \hline
\multicolumn{1}{|c|}{8} &
  Sep 1 - Dec 15 &
   &
   &
   &
   &
  \cellcolor[HTML]{34A853} &
  \cellcolor[HTML]{34A853} &
  \cellcolor[HTML]{34A853} &
  \cellcolor[HTML]{FBBC04} &
  $\sim$50\% &
  $\times$ &
  $\times$ &
  $\checkmark$ &
  $\checkmark$ &
  0.499 &
  0.677 &
  0.662 \\ \hline
\multicolumn{1}{|c|}{9} &
  Oct 1 - Dec 15 &
   &
   &
   &
   &
   &
  \cellcolor[HTML]{34A853} &
  \cellcolor[HTML]{34A853} &
  \cellcolor[HTML]{FBBC04} &
  $\sim$33\% &
  $\times$ &
  $\times$ &
  $<$ &
  $\checkmark$ &
  0.519 &
  0.658 &
  0.643 \\ \hline
\multicolumn{1}{|c|}{10} &
  Aug 1 - Dec 15 &
   &
   &
   &
  \cellcolor[HTML]{34A853} &
  \cellcolor[HTML]{34A853} &
  \cellcolor[HTML]{34A853} &
  \cellcolor[HTML]{34A853} &
  \cellcolor[HTML]{FBBC04} &
  $\sim$60\% &
  $\times$ &
  $\times$ &
  $\checkmark$ &
  $\checkmark$ &
  0.470 &
  0.701 &
  0.683 \\ \hline
\multicolumn{1}{|c|}{11} &
  Jun 1 - Dec 15 &
   &
  \cellcolor[HTML]{34A853} &
  \cellcolor[HTML]{34A853} &
  \cellcolor[HTML]{34A853} &
  \cellcolor[HTML]{34A853} &
  \cellcolor[HTML]{34A853} &
  \cellcolor[HTML]{34A853} &
  \cellcolor[HTML]{FBBC04} &
  $\sim$33\% &
  $>$ &
  $\checkmark$ &
  $\checkmark$ &
  $\checkmark$ &
  \textbf{0.541} &
  0.713 &
  \textbf{0.737} \\ \hline
\multicolumn{1}{|c|}{12} &
  May 1 - Dec 15** &
  \cellcolor[HTML]{34A853} &
  \cellcolor[HTML]{34A853} &
  \cellcolor[HTML]{34A853} &
  \cellcolor[HTML]{34A853} &
  \cellcolor[HTML]{34A853} &
  \cellcolor[HTML]{34A853} &
  \cellcolor[HTML]{34A853} &
  \cellcolor[HTML]{FBBC04} &
  $\sim$90\% &
  $\checkmark$ &
  $\checkmark$ &
  $\checkmark$ &
  $\checkmark$ &
  0.537 &
  0.713 &
  \textbf{0.739} \\ \hline
\end{tabular}%
}
\caption{Effect of temporal range selection on accuracy. Features are generated for different temporal ranges to analyze the effect on outcomes by manipulating the availability of lag and practice features for Presto+ features. AE features are not used, since its temporal range is fixed to a calendar year. Cell colors for months represent how much of the month is covered for that row (blank=none, green=full, orange=half). Values in bold indicate the top-2 performers for each classification task. Nominal sampling frequency of 7 days; *= a sampling frequency of 4 days, **=10 days. $\checkmark$ and $\times$ indicate the availability of a feature type, while $>$ \& $<$ imply that most \& some features available, respectively. Sowing and irrigation feature availability is decided by virtue of which part of the growing cycle the temporal range reveals.}
\label{tab:temp-features}
\end{table*}

\section{Results}\label{sec:results}

\subsection{Dimensional Classification Outperforms Combined Approach}

The dimensional approach--classifying sowing (DSR vs. PTR) and irrigation (AWD vs. CF) separately--significantly outperforms the combined three-class formulation, as shown in~\Cref{tab:results-main}. DSR and AWD achieve F1 scores of 0.796 and 0.742, respectively, supporting our hypothesis that practices are more effectively identified along their distinct dimensions, relative to the 3-class formulations. For those same situations, fusing their outcomes reduces the joint accuracy to $\approx0.6$, confirming that dimensional decomposition avoids error compounding while better reflecting the underlying agronomic processes.

\subsection{Foundation Model Embeddings Enable Robust Classification}
Both Presto and AE yield the highest performance among single feature sources, achieving sowing F1 scores of 0.755 and 0.715, respectively, comparable to the best handcrafted features. Their full value emerges in combination with complementary feature families, particularly Fourier features, where AE+FT attains the best sowing performance (F1=0.798). A similar pattern holds for irrigation, underscoring that foundation model embeddings capture transferable temporal dynamics that enhance orthogonal feature spaces. These results highlight the utility of such embeddings for practitioners, as they are inexpensive to obtain and require no specialized training. Further details are provided in Appendix D. 

\subsection{Lag Features Drive Classification Performance}
\label{sec:temporal-features}

\textbf{Lag features} encode planting-date offsets between plots. The independent PRANA dataset
(Figure A1) shows wide state-level variance but compact practice-specific planting windows, enabling these offsets to act as strong discriminators—though their utility depends on temporal stability across seasons. \textbf{Practice features} capture water-management dynamics intrinsic to each technique. Sowing features (DSR/PTR) are most salient in early growth, while AWD/CF differences emerge later as irrigation frequency diverges (typically every 4–10 days; \Cref{fig:temporal}).

To isolate their effects, we trained models on restricted temporal windows (\Cref{tab:temp-features}). Sowing models perform best when both lag and early-season features are available (rows 1, 3, 10–11), confirming that they rely on both signal types and remain robust to sampling frequency (4–7 days). In contrast, irrigation models depend almost entirely on lag information: performance peaks when all features are present (row 11) and collapses once lag cues are reduced (row 8) or sowing phases are absent (rows 5–9). This dependence indicates that AWD detection leverages schedule shifts rather than true irrigation cycles, which Sentinel-1A’s 12-day revisit cannot resolve. Further analyses of feature importance, orbit effects, and errors are provided in Appendix F.

\subsection{Inference at Scale}
\label{sec:inference-scale}

A key challenge in promoting sustainable agriculture is understanding adoption patterns of water-saving practices. Despite extensive farmer training, PRANA lacked state-wide visibility due to the high cost of surveys. Our satellite-based approach overcomes this limitation, enabling monitoring of Punjab’s 3.2 million hectares of rice cultivation and providing critical insights for food and water policy. We delineate field boundaries using the FTW algorithm~\cite{kerner2024fields} applied to bi-temporal Sentinel-2 RGB-NIR imagery (May \& Oct 2024). FTW polygons are filtered to 0.2-10 Ha, matching Punjab’s typical rice-plot size—means: 0.5±0.7 Ha (PRANA internal survey), 0.6±0.5 Ha (~\citep{wang2022unlocking} Punjab subset), and 2.7 Ha \citep{singhEconomicsBasmatiRice2024}. Validation of FTW polygons over the PRANA training area shows that 57\% of sub-1 Ha plots were detected, with mIoU 0.41 ±0.2 over all plots. FTW often splits large fields ($\approx$1.4 sub-plots per label) and misses 36\% of plots, reflecting both its resolution limits and ability to capture smaller sub-plots. We provide further details and qualitative samples of this step in Appendix G and Figure A9.

Given sparse training labels ($\sim$1,400 labeled plots vs. 3 million during inference), we employed an ensemble of top-performing RF and GB models, aggregating predictions via the modal class. Results are summarizes in~\Cref{fig:infer2}A, showing marked spatial heterogeneity in DSR adoption across and within districts. Our district-level predictions correlate strongly with government records~\cite{progressivefarming2025} reporting 253,328 acres under DSR (Spearman’s $\rho$=0.69, RBO=0.77~\cite{webber2010similarity}), despite regional discrepancies. Since our training data included only rice plots, we applied a rice-cultivation mask~\cite{han2022annual} from 2021 to exclude non-rice fields. We observe a residual diagonal gradient arises from minor temporal offsets between adjoining S-1 rows, mitigated via radiometric terrain-geocoded correction.

The model generally overestimates DSR area relative to government figures, except in the top three districts. These discrepancies may stem from both model error and noise in the self-reported government data\footnote{Data sourced from a DSR incentive program offering Rs. 1500 per acre, pre-verification; c.f. see ~\cite{progressivefarming2025} Table 1}, which, as seen in neighboring states~\cite{Deswal2024DSRFake}, may include unverifiable claims. Notably, for Sri Muktsar Sahib and Fazilka, masking introduced large differences ($\sim$40k acres each), likely due to post-2021 expansion in rice cultivation. We therefore report unmasked estimates for these districts (see Table A1 and Appendix G).
Further in-field validation will be important to confirm these large-scale adoption estimates. Despite such variations, the high RBO score indicates strong preservation of district rankings, making the system well-suited for policy prioritization by highlighting both lagging and leading districts for targeted action and learning best practices.

\section{Limitations}
\label{sec:limitation}

Our approach, while effective, faces several constraints that merit consideration. First, the current S1 revisit frequency of 12 days limits our ability to capture high-frequency irrigation variability characteristic of AWD practices. This temporal resolution challenge could be addressed by integrating data from ascending and descending orbits or incorporating forthcoming SAR missions.

Second, field boundary extraction using the FTW algorithm is constrained by Sentinel-2's spatial resolution, which imposes a minimum detectable plot size. This limitation is particularly salient in smallholder farming contexts where field sizes may fall below this detection threshold, potentially excluding smaller agricultural parcels from analysis.

Third, explainability analysis revealed that our models have degrees of reliance on temporal lag features that capture distinctions between water management practices. Additionally, the small dataset size ($\sim$1,400 plots) relative to the large inference dataset underscores the need for careful validation. Although Punjab has relatively low diversity in its agroecological zones, other regions may present greater variation in cropping systems, climate, or practice clustering, affecting generalization. Deployment of such methods would require systematic post-training field validation to verify model predictions against ground truth before operational use, particularly given our reliance on temporal patterns that may shift across agricultural contexts. Therefore, cross-year and cross-region generalization remain future extensions to assess temporal and spatial transferability.

Finally, our analysis depends on external estimates of rice-growing areas. Temporal mismatches between these masks and the study period introduced complications, particularly in regions where rice cultivation has expanded. Future work should explore explicitly classifying non-rice plots to reduce reliance on exteral and outdated maps.

\section{Conclusion \& Future Work}

Our study demonstrates the feasibility of satellite-based monitoring for rice water management, offering a scalable and data-driven framework for tracking sustainable agriculture adoption. By decoupling classification into sowing (DSR) and irrigation (AWD) dimensions, the approach outperforms combined formulations. Co-designed with agronomy experts, this dimensional structure proved essential: DSR detection notably surpassed joint classification and enabled statewide mapping of DSR adoption across 3 million fields in Punjab, India. Explainability analysis further showed that while AWD classification achieves an relatively high F1 scores, it depends primarily on differential planting schedules rather than irrigation signals, underscoring the importance of interpretability for reliable deployment. AWD performance remains limited by Sentinel-1A’s 12-day revisit, though upcoming missions (Sentinel-1C/D, NISAR) will improve temporal resolution, potentially improving irrigation detection.

Operating without prior knowledge of planting dates, the framework generalizes to real-world systems with heterogeneous cropping calendars. Embeddings from remote sensing foundation models--Presto~\cite{tseng2023lightweight} and AlphaEarth~\cite{brown2025alphaearth}--were key to this generalization, delivering strong accuracy from limited labeled data. A new class of efficient and highly performant EO foundation models, including OlmoEarth\footnote{\url{olmoearth.allenai.org}}, Panopticon~\cite{waldmann2025panopticon} and Galileo~\cite{tseng2025galileo}, have recently emerged, with precomputed embeddings increasingly available, further extending the potential for scalable, low-cost monitoring of agricultural practices.
Finally, future work should examine why AlphaEarth’s year-long embeddings retain strong sensitivity to rice sowing signals despite spanning multiple cropping seasons.

District-level predictions correlate strongly with government records, supporting utility for food and water policy through: (1) evidence-based resource allocation, (2) large-scale impact measurement, (3) identification of low-adoption areas, and (4) statewide monitoring without costly field surveys. At finer scales, plot-level irrigation mapping provides timely, high-resolution data on water management practices, enabling biogeochemical models to more accurately simulate methane emissions and quantify the mitigation benefits of sustainable agriculture. Extending this framework to other rice-growing regions and adapting it to local conditions--crop varieties, irrigation regimes, and management practices--will enhance generalization and policy relevance. As climate change accelerates water scarcity, scalable monitoring of sustainable irrigation will be vital for ensuring food security and advancing the Zero Hunger Sustainable Development Goal.

\newpage

\bibliography{aaai2026}

\clearpage

\appendix
\appendix
\renewcommand{\thefigure}{A\arabic{figure}}
\renewcommand{\thetable}{A\arabic{table}}
\setcounter{figure}{0} 
\setcounter{table}{0}  

\section*{Appendix} 

\section{Data and Code Availability}
The data and code that enables study replication shall be made available at the end of the peer-review period.

\section{PRANA Program}

We focused our analysis on the Indian state of Punjab, both because of the scale of rice farming, and importance to national and global rice supply, as well as the existing program by The Nature Conservancy, the Promoting Regenerative and No-burn Agriculture (PRANA) project, to promote water-saving rice practices as well and reduce crop residue burning. India is the world's second biggest producer of rice ~\cite{AgriculturalProductionStatistics}, with the state of Punjab producing 10.42\% of the country's rice in 2024 at 144 million tons~\cite{MARKETINTELLIGENCEREPORT2024} representing roughly 3.7\% of the world's rice production capacity. Punjab's prominence in rice production is heavily supported by irrigation infrastructure and high-yield farming practices but raises concerns about water sustainability due to the heavy reliance on groundwater systems. In Punjab, the adoption of the rice-wheat cropping system has led to a manifold increase in the irrigation water demand, and about 73\% of it is met from groundwater~\cite{sidhu2021spatio}, with the state experiencing a strong decline groundwater capacity~\cite{rodell2009satellite} in the last few decades. A large part of this is due to the demand for water for rice cultivation, since the state of Punjab is in a semi-arid region with highly seasonal monsoon precipitation. Thus, unreliable surface water supplies coupled with excessive groundwater exploitation due to free electricity~\cite{gupta2023free} has led to a long-term groundwater decline of 41.6 cm/yr in the state~\cite{baweja2017groundwater}. Han et al. ~\cite{han2022annual} find that rice paddy planting regions increased significantly in northwestern India - one of the regions with the highest conversion from single- to double-cropping of rice.

\begin{figure*}[!htbp]
    \centering
    \includegraphics[width=0.95\textwidth]{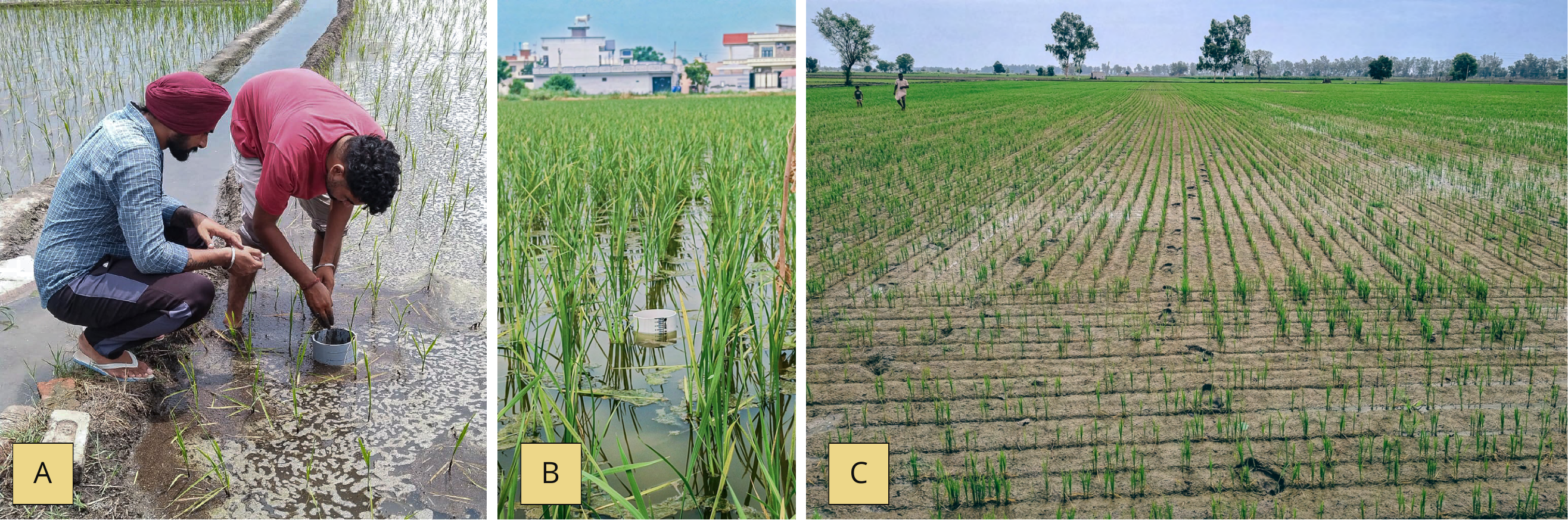}
    \caption{Water-saving Interventions: A) AWD measuring pipe being installed in a demonstration field. B) AWD pipe after installation. C) Photograph of a sample field practicing DSR taken during the tillering stage; note the lack of flooding in the field.}
    \label{fig:sup:practices}
\end{figure*}

The PRANA program conducted focused training sessions for \textasciitilde150,000 farmers on a targeted scale to generate awareness, provide education, agronomic support and monitoring of farmers' irrigation practices over the years 2023-2024, to enable adoption of both DSR and AWD methods. For both DSR and AWD, potential areas were identified based on the soil type and availability of required machinery. Field demonstrations was conducted in nearly every five villages to showcase the techniques locally. For AWD practice, 27,011 pipes were provided to the farmers pro bono and 26,997 hectares were covered under AWD using these pipes. Apart from these activities, paddy crop guides on agronomic practices were designed and provided to the farmers in local languages to build and strengthen adoption capacity. Figure~\ref{fig:sup:practices} shows field photos where these interventions were put into place in 2024.

\subsection{Dataset}

The dataset used in this study was collected by the authors in collaboration with the PRANA program. Due to the sensitive nature of exact geo-coordinates, and the challenges associated with the traditional method of de-identification (jittering will cause major data fidelity issues), we have chosen to scrub the geo-coordinates from the dataset. Unfortunately, this implies that users wishing to recreate the results will not be able to use the source data (Sentinel-1) to reproduce the steps. Therefore, we provide all the features created for each plot including hand-crafted (HC), Presto (P), Fourtier Transform (FT) and Google AlphaEarth embeddings (AE). These have been packaged with the codebase. Additionally, we provide all code and steps needed for users to generates features for any given geo-referenced polygons.

\subsection{Prediction in the Presence of Planting Date Heterogeneity}

Previous studies that discriminated DSR from PTR relied on synchronizing SAR backscatter data with cropping calendars, enabling direct modeling of temporal signatures~\cite{gumma2015mapping, fikriyahDiscriminatingTransplantedDirect2019, villano2019separability}. While effective, this approach is inherently limited in scalability, particularly in regions with unknown or heterogeneous planting dates. 

\begin{figure}[ht]
    \centering
    \includegraphics[width=\linewidth]{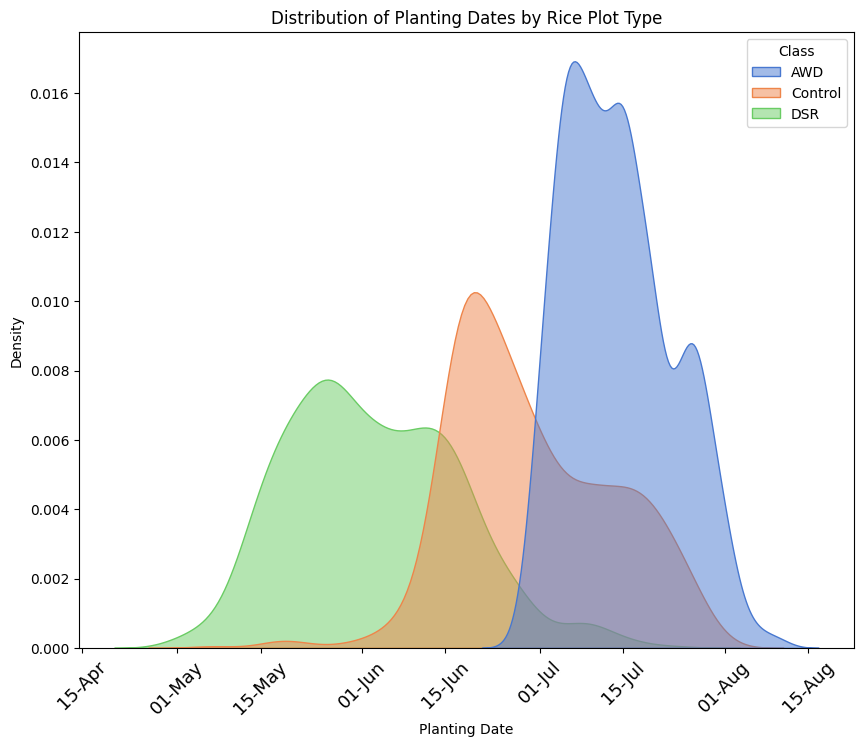}
    \caption{Planting date distributions across water management practices of interest.}
    \label{fig:sup:pd_dist}
\end{figure}

As shown in Fig. \ref{fig:sup:pd_dist}, the planting dates in our dataset exhibit a roughly trimodal distribution: DSR plots are sown first, followed by control plots, and finally AWD plots. This temporal disparity spans up to 110 days between the first and last planting dates, introducing substantial temporal variability both between and within classes. This heterogeneous distribution naturally introduces temporal differences between classes, and where the distributions are wide enough - within classes. By not synchronizing cropping calenders across sample plots, we introduce \textit{lagged features}—temporal offsets capturing the relative differences in planting dates—into our analysis. Lagged features enable the model to account for inter-class and intra-class variability in planting schedules, providing discriminatory power without requiring explicit cropping calendar data. Therefore lagged features can act as a proxy for cropping calenders as long as they exhibit low dispersion within a class. In practice, however, these features also reflect the limitations imposed by wide planting date distributions, which can span up to 2.5 months within individual classes like DSR and PTR as seen in Fig. \ref{fig:sup:pd_dist}.

Although planting and harvesting dates were available in our dataset, we intentionally excluded them from the modeling process to enhance the generalizability of our framework. This decision ensures that our approach remains applicable to regions where such information is unavailable. Additionally, farmer interviews revealed the presence of "third crops" (such as lentils, maize, or green manuring crops\footnote{a green manure is a crop specifically cultivated to be incorporated into the soil while still green}) planted just before the kharif rice season. These crops introduce further variability in SAR backscatter signatures, complicating the detection of water management practices. Due to incomplete data on these practices, third-crop effects were not explicitly modeled but represent an additional source of temporal heterogeneity.

\begin{figure*}[!htbp]
   \centering
   \includegraphics[width=0.9\textwidth]{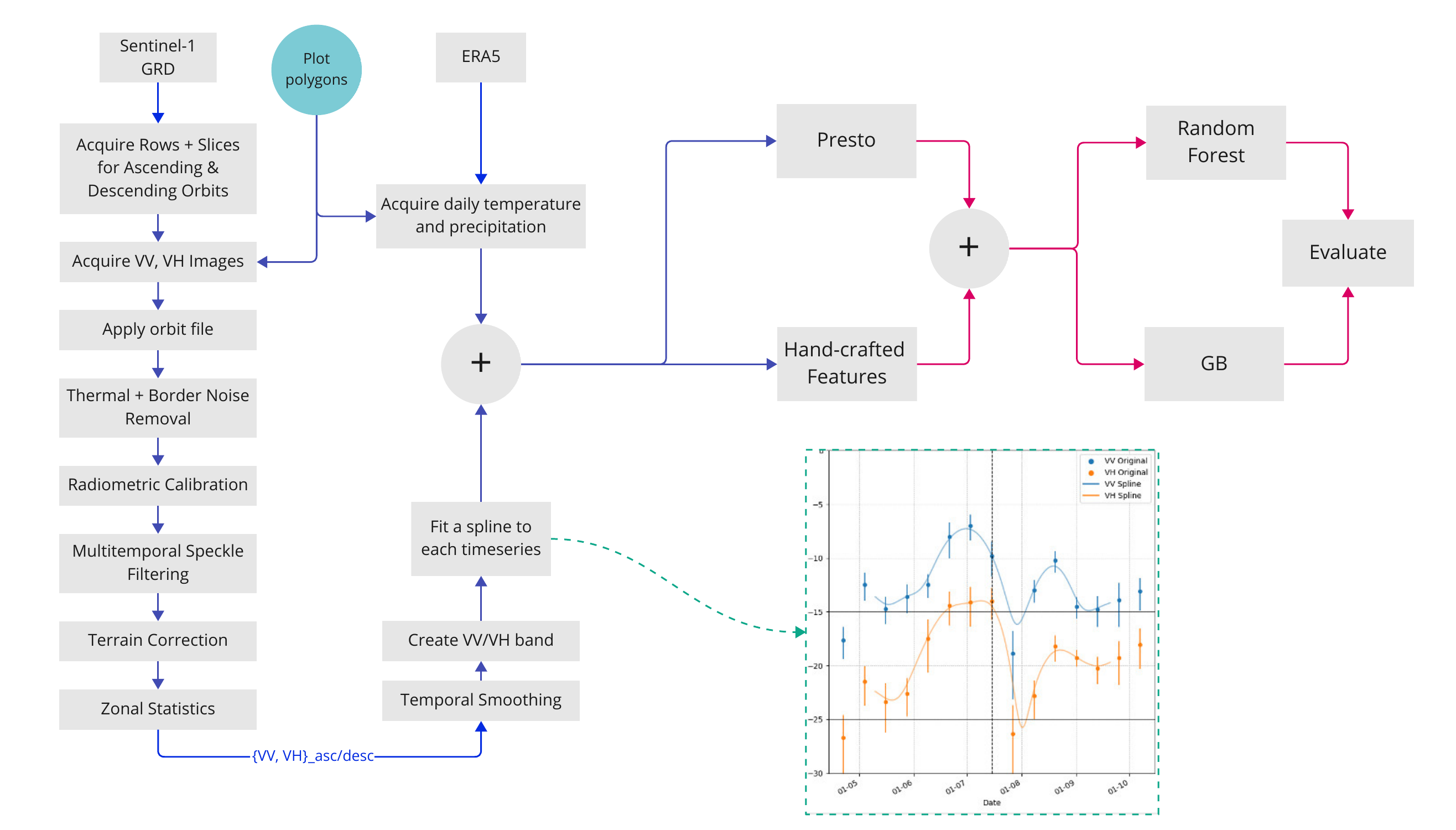}
   \caption{Training and validation workflow used in the study. We utilize Copernicus Dataspace and GEE to acquire Sentinel-1 and ERA5 temporal data on a per plot basis, and to subsequently reduce them to plot level statistics. We then fit a spline, resample, generate hand-crafted features and embeddings from Presto over a given time period. The separability of water management classes is modeled using a Random Forest and Gradient Boosted Trees. The inset shows an example of the average S-1 VV and VH timeseries after smoothing and spline fitting, with confidence intervals at each point in time.}
   \label{fig:sup:workflow}
\end{figure*}

\section{Modeling Details}
\label{sec:sup:modeling}

Sentinel-1 timeseries is generated by the process explained in \cref{sec:sat-data}. Additionally, for all features, we experiment with resampling the smoothed spline at frequencies of 4, 7 and 10 days. This allows us to modulate the temporal resolution of the data where possible (when the total number of timesteps $\leq$ 24) or to accommodate longer temporal ranges (such as rows 11-12 in \Cref{tab:temp-features}).

We experiment with ensemble-based learning methods including random forest (RF) and gradient-boosted trees (GB) to classify rice plots into one of 3 possible classes: AWD, DSR or PTR. Previous works have shown the suitability of these methods to LULC and agricultural practice mappings ~\cite{liuUsingSentinel1Sentinel22022, fangComprehensiveReviewRice2024a, rudiyantoAutomatedNearRealTimeMapping2019, TATSUMI2015171}. Since we can reduce plot-level features to a point estimate, time-series modeling is possible with ensemble models like RF or GB. We use scikit-learn's implementation for RF and LightGBM for GBs. Additionally we employ Dynamic Time Warping along with kNN classification as baseline measure of performance, that directly compares each test timeseries to the training datasets to infer class.\\

\subsection{Hand-crafted Features} 
\begin{figure*}[!ht]
    \centering
    \includegraphics[width=1\textwidth]{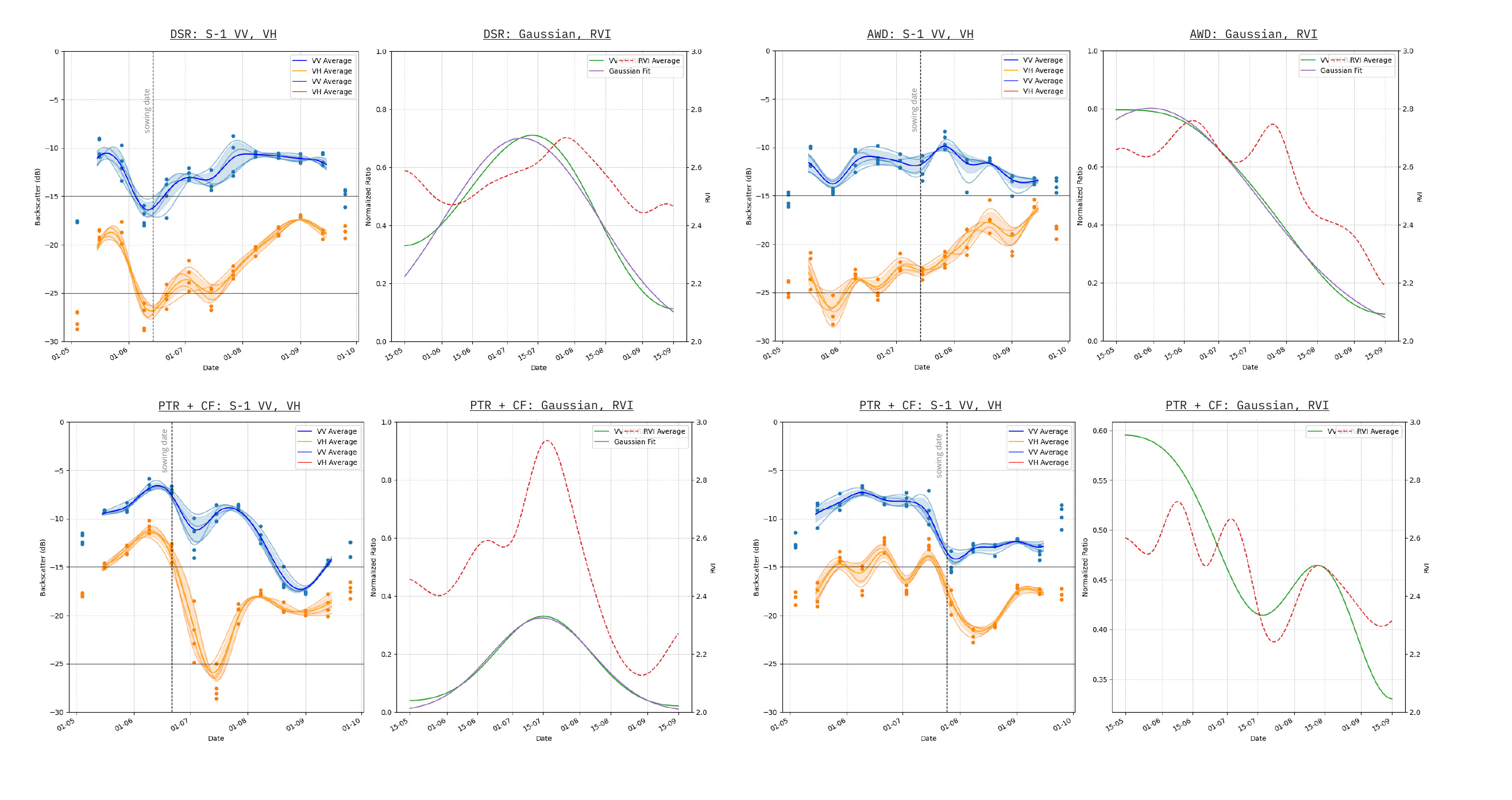}
    \caption{Examples of hand-crafted features generated from SAR backscatter timeseries for different irrigation field types. Top-left: DSR and CF. Top-right: PTR and AWD. Bottom row: PTR with CF. Left subplots show the VV (blue) and VH (orange) timeseries for every pixel contained within a plot, and the dark blue and dark orange show the mean (with shaded regions denoting the standard deviation) VV and VH, respectively, per plot. Right subplots show the gaussian fit to the VV/VH curve and the radar vegetation index (RVI). The dotted lines in each left-subplot denote the planting date for that plot.}
    \label{fig:supp-s1-feats}
\end{figure*}

We generate a set of hand-crafted features from each S-1 band ($VV, VH, VV/VH$) per orbit, as the summary statistics for a rice-growing plot. For each band, we locate the positions of the first $K$ troughs and crests as well as the first $K$ inflection points. We extract the timestamp and amplitude at these points, measuring timestamp relative to the first date in the timeseries, in number of days. Experimentally, it was found that $K=3$ would cover all timeseries in the training dataset, with larger $K$s leading to missing values. We also featurize the total number of trough and crests, along with the mean, minimum and maximum amplitude values of each band per orbit. Following previous work~\cite{bazziMappingPaddyRice2019}, we fit a Gaussian kernel to the VV/VH ratio and employ its parameters (amplitude, day of peak, standard deviation and $R^2$ of kernel fit) as additional features. Additionally, we calculate the Radar Vegetation Index (RVI) as modified for S-1~\cite{nasirzadehdizaji2019sensitivity} as it has been shown to be useful in detecting crop height. The timeseries generation process for S-1 bands after processing and additional hand-crafted features are visualized in Fig. ~\ref{fig:supp-s1-feats}.

\subsection{Presto Features} 
In addition to hand-crafted features, we employ the Pretrained Remote Sensing Transformer (Presto)~\cite{tseng2023lightweight} as a temporal feature extractor. Presto is transformer-based deep learning model that has been pre-trained on remote sensing pixel-timeseries data, including Sentinel-1 SAR, Sentinel-2 multispectral, Dynamic Earth land use and land cover (LULC) labels, ERA5 temperature and precipitation datasets and, latitude \& longitude. To extract embeddings, we freeze the Presto encoder, fix the Dynamic Earth class to `cropland', and input both ERA5 variables and Sentinel-1 VV and VH bands from either ascending or descending orbits. Since we fit a spline to smoothed SAR timeseries, we can use arbitrary sampling frequencies to pseudo-assess the effect of sampling frequency. This is especially important as a limitation of the Presto encoder is that it cannot process more than 24 timesteps-in such cases we can increase the sampling duration to fit the entire timeseries. Therefore we choose $f$=4,7,10 days. Additionally, we extract embeddings from the encoder at each timestep \footnote{Presto takes a timestamp numbering for the month number, however we find that it gives better performance to use the week number instead of the month, since irrigation and sowing signal frequency is higher than a month}, plus one feature for the location, generating a total of $128 \times (T\times2)\footnote{One set of embeddings for S1 and Dynamic World inputs} + 1$ embeddings. This process is encapsulated within the `Presto' block of Fig.~\ref{fig:sup:workflow}.

\subsection{AlphaEarth Embeddings}
\label{sup:ae}

AlphaEarth Foundations~\citep{brown2025alphaearth} generates geospatial embeddings through a spatiotemporal encoder–decoder architecture that jointly models multi-source Earth observation inputs across continuous time. Each embedding represents a 64-byte vector summarizing environmental and surface dynamics over a defined ``valid period'', learned from heterogeneous inputs (e.g., Sentinel-1/2, Landsat, GEDI, ERA5-Land, GRACE, NLCD, and Wikipedia text). The model’s Space-Time Precision (STP) encoder integrates spatial self-attention, temporal axial attention, and convolutional “precision” operators with Laplacian pyramid rescaling to preserve both global context and local spatial detail. An implicit decoder reconstructs sensor-specific variables from the embeddings using conditional time and sensor metadata, regularized by a batch-uniformity loss enforcing uniform distribution on the unit hypersphere $S^{63}$. Training employs contrastive learning between teacher–student video models and a text-alignment branch, producing dense, continuous 10 m embeddings that encode the joint spatiotemporal structure of the Earth system, which is what we use as the AE product, available from Google Earth Engine at this link: \url{https://developers.google.com/earth-engine/datasets/catalog/GOOGLE_SATELLITE_EMBEDDING_V1_ANNUAL}.

\subsection{Modeling Details}

For RF, we sweep over the following ranges for each hyperparameter:
\begin{itemize}
    \item n\_estimators: [300, 1800]
    \item max\_depth: [2, 20]
    \item min\_samples\_split: [2, 20]
    \item min\_samples\_leaf: [1,20]

\end{itemize}
For GB, we sweep over the following ranges for each hyperparameter:
\begin{itemize}
    \item max\_depth: [3,8]
    \item learning\_rate [0.001, 0.01]
    \item n\_estimators: [100, 500]
    \item subsample: [0.6,1.]
    \item colsample\_bytree: [0.6, 1.]
    \item min\_child\_samples: [5, 30]
    \item min\_split\_gain: [1e-8, 0.1]
    \item num\_leaves: [20, 3000]
    \item reg\_alpha: [1e-8, 0.5]
    \item reg\_lambda: [1e-8, 0.5]
\end{itemize}

All model details can be found in the code file under scripts/train\_model.py\\

For all models and feature combinations, except the random baseline, we run a single training run with a seed=42. The random baseline is run 1000 times and the mean±stdevs are reported in ~\Cref{tab:results-main}.

All model training was done on a x86\_64 Linux machine with 250Gb of RAM and 64 CPUs. A L40s GPU was used to generate the Presto embeddings, but this can be achieved on any CPU. All software versions are available in requirements.txt file.

Final model weights for the best DSR and AWD models are also provided in data/models.

\begin{figure*}[t]
    \begin{minipage}[t]{0.48\textwidth}
        \centering
       \includegraphics[width=\linewidth]{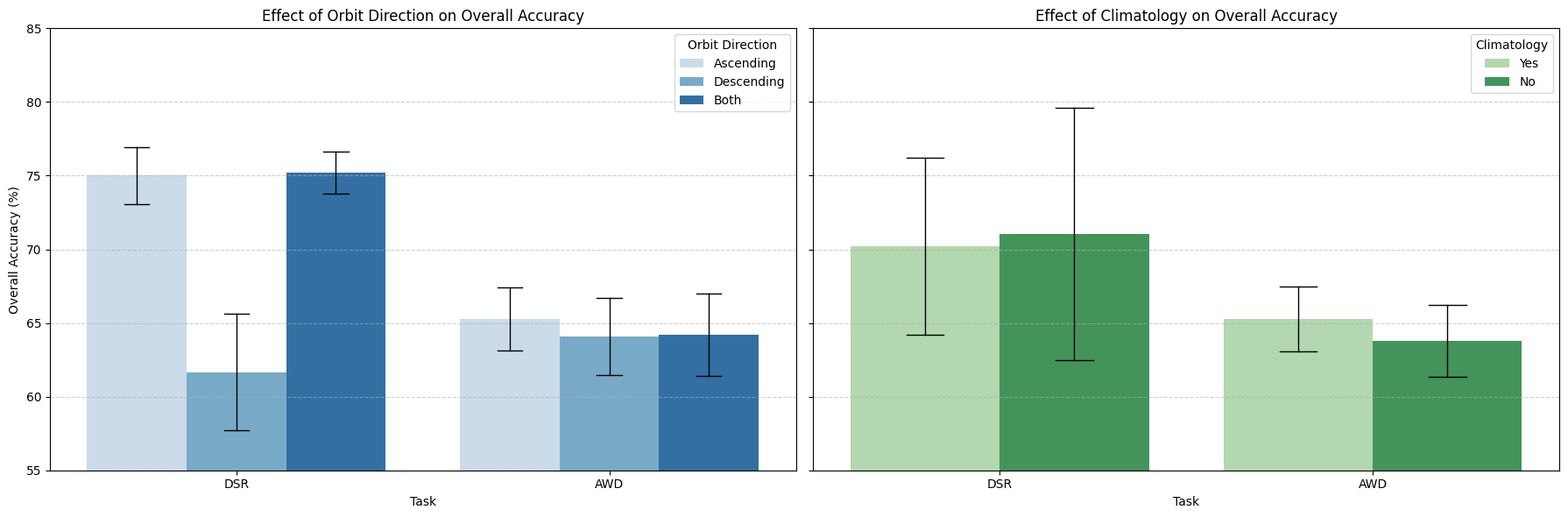}
       \caption{Impact of orbit direction and climatology data (temperature and precipitation) on overall accuracy of DSR and AWD classification performance. Left: Bar plot grouped by orbit direction (ascending, descending, both) with confidence intervals highlighting variability within each task. Right: Bar plot showing the impact of climatology inclusion (yes, no) on OA, with confidence intervals highlighting variability within each task.}
       \label{fig:sup-orbit-clim-fs}
    \end{minipage}
    \hfill
    \begin{minipage}[t]{0.48\textwidth}
       \centering
       \includegraphics[width=\linewidth]{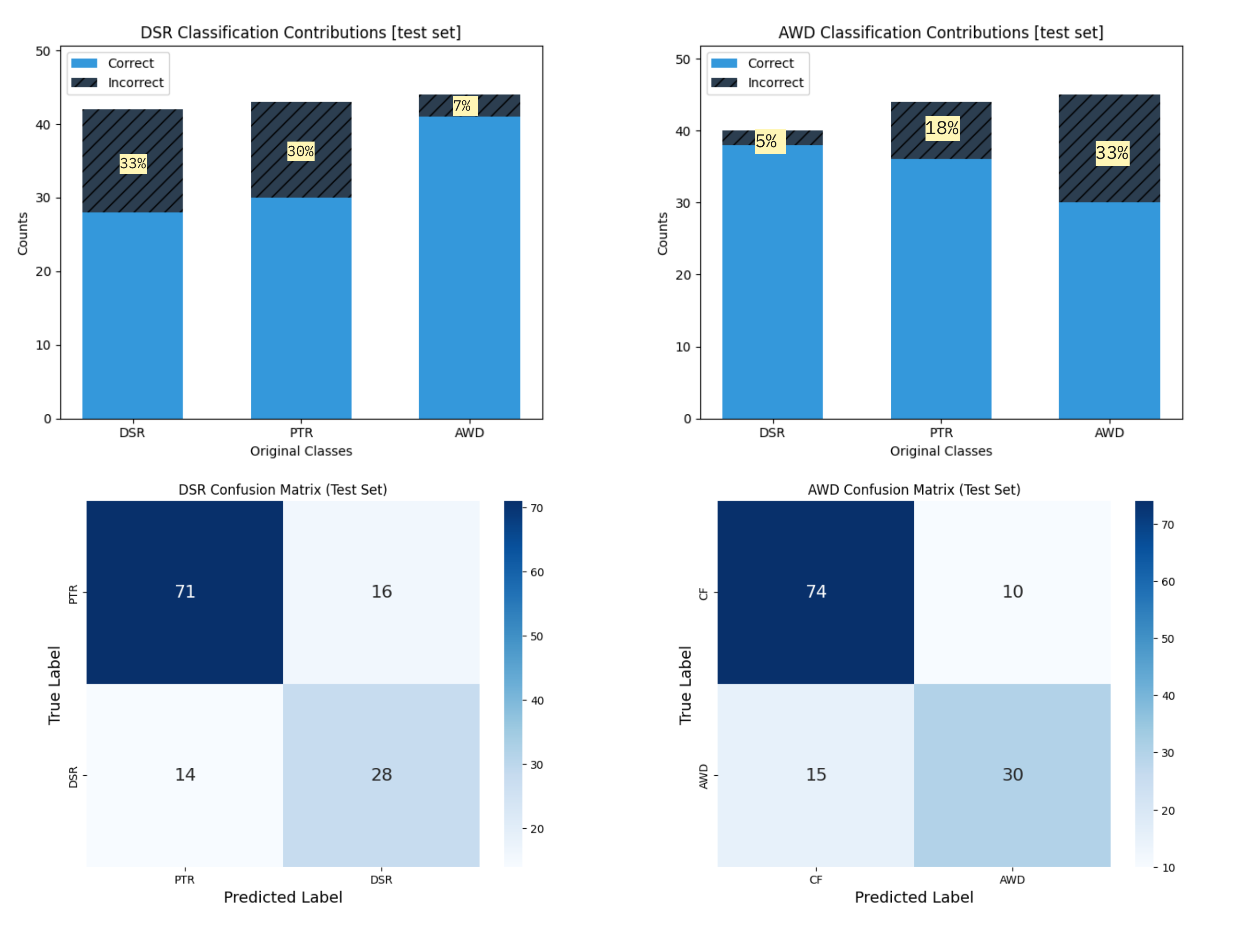}
       \caption{Error Analysis. Top-left: DSR classification error percentages on test set from original classes. Bottom-left: DSR classification confusion matrix from collapsed classes. Top-right: AWD classification error percentages on test set from original classes. Bottom-right: AWD classification confusion matrix from collapsed classes. Text on bars represents percentage of errors contributed from that class.}
       \label{fig:error-dsr-awd}
    \end{minipage}
\end{figure*}

\section{Foundation Model Performance}
\label{sec:sup:fm}

From \cref{tab:results-main}, it is evident that features derived from remote-sensing foundation models--Presto and AlphaEarth (AE)--consistently dominate performance, particularly when combined with complementary feature families such as Fourier (FT) and handcrafted (HC) descriptors. While HC features require domain expertise in SAR signal behavior and agronomic processes to construct, FT and DTW are standard numerical transformations applicable to most temporal datasets with minimal preprocessing.

The strong performance of FT–FM combinations underscores a key finding: foundation model embeddings capture rich temporal representations that align well with frequency-domain features, jointly enhancing separability of both sowing and irrigation practices. This synergy suggests a practical route for operational deployments, especially for practitioners lacking extensive SAR expertise.

Among the two models, AE offers the most accessible path to adoption, as its embeddings are freely available on Google Earth Engine for research and non-profit use. In contrast, Presto requires local preprocessing of Sentinel-1 data (noise removal, calibration, speckle filtering, terrain correction) before embeddings can be generated, though it remains valuable when high temporal fidelity is required. Together, these models highlight the growing utility of pretrained Earth-observation representations for agricultural monitoring: practitioners can directly extract embeddings, pair them with simple numerical transforms such as FT, and train lightweight classifiers for diverse crop-mapping tasks without retraining large models or acquiring new ground data.

Surprisingly, AE embeddings—which summarize temporal dynamics over a full calendar year—achieved the highest performance, despite rice cultivation in Punjab occurring only between April and December and varying by several months across plots. This suggests that AE captures persistent spectral–textural cues linked to sowing and irrigation transitions, even when embedded within multi-crop annual cycles. Further investigation is needed to understand why features derived from year-long multisensor inputs remain sensitive to rice-specific signals without explicit temporal masking, unlike Presto, for which we constrained inputs to the active growing season.


\section{Ablations}

\subsection{Effect of Sentinel-1 Orbit} 
For DSR classification, ascending orbits consistently outperformed descending orbits, as shown in Fig.~\ref{fig:sup-orbit-clim-fs} (left). This suggests that radiometric properties of the descending orbit make it suboptimal for capturing DSR features in this region. The use of both orbits showed no significant advantage over ascending orbit alone, and the additional effort required to process both orbits may not be justified. For AWD classification, orbit direction had less impact, likely due to lag features driving performance regardless of orbit, while AWD-specific features were not captured effectively. This highlights that ascending orbits better capture DSR features compared to descending orbits in this study.

\subsection{Effect of Climatology Data} 
High overlap in CIs for models with and without climatology data indicates no significant benefit from including these variables (Fig.~\ref{fig:sup-orbit-clim-fs} , right). As the average performance without ERA5-Land data was slightly higher, and considering the effort required to acquire and integrate this dataset, there appears to be no inherent advantage to its inclusion for this region and task.

These findings suggest that consistent use of ascending orbits is optimal for DSR classification, while climatology data may be omitted without loss of performance, simplifying large-scale inference.

\section{Error Analysis}
\label{sec:sup:error}

Given the distributed but nearly separable cropping calendar (Fig.~\ref{fig:sup:pd_dist}), it is useful to analyze the sources of classification errors with respect to the original classes. For the best-performing DSR classification model, Fig.~\ref{fig:error-dsr-awd} (top-left) highlights the proportion of test set errors originating from each original class, showing a linear decrease in errors as planting dates increment from left to right. Similarly,  Fig.~\ref{fig:error-dsr-awd} (top-right) illustrates the proportion of incorrect samples and their original classes for AWD classification, with errors decreasing towards the left, for dates further from the AWD class.

For both tasks, errors predominantly stem from temporally similar samples ($\sim$33\% of that class if incorrectly classified), suggesting that lag features may not provide sufficient discrimination in these cases and that practice features lack the necessary power. Confusion matrices (Fig.~\ref{fig:error-dsr-awd} bottom row) reveal that negative class scores dominate the results. For both classification tasks, the positive classes(DSR, AWD) achieves precision of 75\% and recall of 67\%.

\paragraph{Importance of orbit direction:}\label{sec:optimizing}
For operational monitoring at scale, we found ascending orbit data consistently outperforms descending orbit data for DSR classification, with no benefit from combining orbits or including climatology variables (\Cref{fig:sup-orbit-clim-fs}). This streamlined approach reduces computational requirements while maintaining performance during inference.

\subsection{Feature Importance}

\begin{figure*}[!ht]
   \centering
   \includegraphics[width=\textwidth]{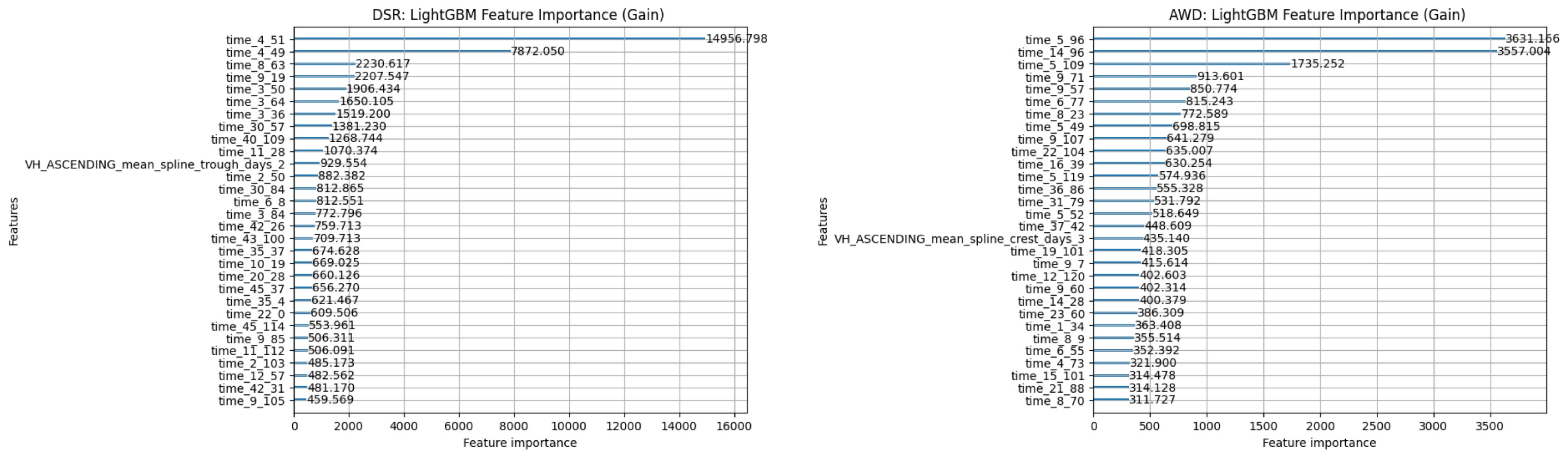}
   \caption{Feature importance: Using the gain attributed to each feature, the feature importance for LightGBM models are estimated for sowing (DSR) classification, left and irrigation (AWD) classification on the right.}
   \label{fig:feature-imp}
\end{figure*}

For the best performing models for DSR and AWD classification, we estimate the feature importance as shown in Fig.~\ref{fig:feature-imp}. For DSR classification, the best performing model was trained on the time period Jun 1- Sep 5 with a 4 day sampling frequency (Fig.~\ref{fig:temporal}, row 3). Therefore the features that start with ``time\_*'' refer to the time period in 4 day increments beginning on Jun 1\footnote{Since we pass in both S1 timesteps and a single fixed value for Dynamic World, the Presto Encoder generates: [0] lat-lon token, [1:T+1] temporal tokens, [T+1:2T+1] Dynamic World tokens}. The top two features are therefore from the time period Jun 16-20, implying that the features in those regions were most useful in classification. From Fig.~\ref{fig:sup:pd_dist}, using the lens of sowing detection, we see that this refers to the time when the majority of PTR crops have been sowed (control/DSR has peaked and AWD has begun), which aligns with our conclusion that the sowing features are actually helpful.

For AWD classification the model was trained for the period May 1 to Dec 15 with a ten day sampling frequency (Fig.~\ref{fig:temporal}, row 12). Similarly, the top two features (Fig.~\ref{fig:feature-imp} left) correspond to the time periods of June 20-30, and Sep 18-28. The former is roughly the same time-period as the top-2 features for DSR, while the latter likely captures some important lag feature, since all plots are at some stage of growth in September, regardless of when they were planted.

\begin{figure*}[!htbp]
   \centering
   \includegraphics[width=\textwidth]{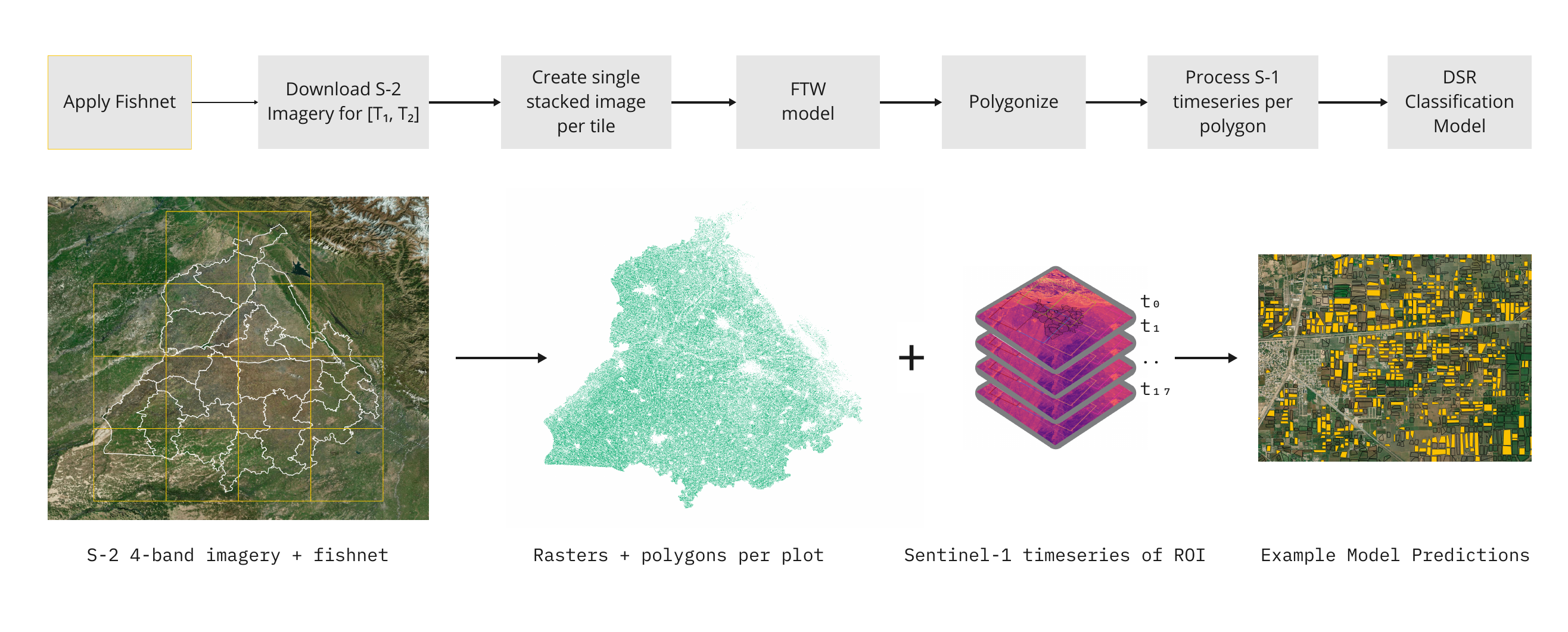}
   \caption{Inference Workflow: Methodology followed during DSR classification predictions for 3 million polygons using S-1 timeseries. FTW~\cite{kerner2024fields} polygons are generated from bitemporal 4-bands S-2 imagery}
   \label{fig:sup:scaling-workflow}
\end{figure*}

\paragraph{Impact of temporal overlap:}
For both DSR and AWD classification, errors occur predominantly in samples with temporally similar planting dates (\Cref{fig:error-dsr-awd}). This error pattern confirms that while our dimensional approach effectively addresses most planting date heterogeneity, classification remains challenging when practices exhibit similar temporal signatures. The elevated error rate in AWD classification in the presence of irrigation features shown in \Cref{tab:temp-features} rows 8-10 further corroborates our finding that current satellite revisit frequencies inadequately capture irrigation cycles. These rows were additionally modeled with a higher sampling frequency of 4 days, yielding even poorer results, showing that higher sampling frequency has no effect on AWD performance.

\section{Large-scale Inference}
\label{sec:sup:inference}
An important goal of this study is to understand the spatial extent of use of water-saving rice management practices in Punjab. Out of the 150,000 farmers who received training and support for the 2024 kharif season, it was infeasible to track the practices of all participants and their networks during the season to measure the efficacy of these interventions. The data collection efforts of PRANA were limited to only a small fraction of the potential market. Here we present our findings on scaling the model's predictions across the state at a plot-level.

The entire workflow for large-scale inference is outlined in Fig.~\ref{fig:sup:scaling-workflow}. In order to make predictions at a plot-level, we first need to delineate agricultural plots across the state. There are no current estimates of the number of plots, although the total rice-growing area in 2021 was roughly 3.2 million hectares (ha) as surmised from Han et al.~\cite{han2022annual}. Given that average plot sizes range from 0.5 to 2 ha, delineation of millions of plots would be required, making manual efforts impossible. We apply the automated method known as Fields of the World (FTW)~\cite{kerner2024fields} to delineate approximately 3 million fields across the state. The FTW model is a U-net with EﬃcientNet-b3 backbone trained on concatenated bitemporal 4-channel RGBN images (with and without green cover, typically a few months separated), which are then polygonized to compute object-level metrics using an IoU threshold of 0.5.

We acquire Sentinel-2 (S-2) 4-band\footnote{FTW only requires RGB-NIR bands} bi-temporal images from the months of May and October 2024, representing cloud free periods representing windows A \& B respectively, from GEE. However, due to GEE restrictions on the amount of downloadable imagery, the region of interest needs to be split into non-overlapping tiles by applying a `fishnet' and downloading the respective tiles. We merge the tiles and stack the bi-temporal images as a single 8-band image (Fig \ref{fig:sup:scaling-workflow}). This image is fed as an input to the FTW model which produces a 3-band raster\footnote{The bands are: fields, non-fields and boundaries: we keep only the field band.}, which we subsequently convert to polygons, with a minimum area of 2000$m^2$. We also filter out any polygons larger than 10 ha. to reduce spurious polygons that are unlikely to be rice growing plots, given the nature of smallholder agriculture in the state.

\begin{figure}[!t]
    \centering
    \begin{subfigure}[b]{\linewidth}
        \centering
        \includegraphics[width=\linewidth]{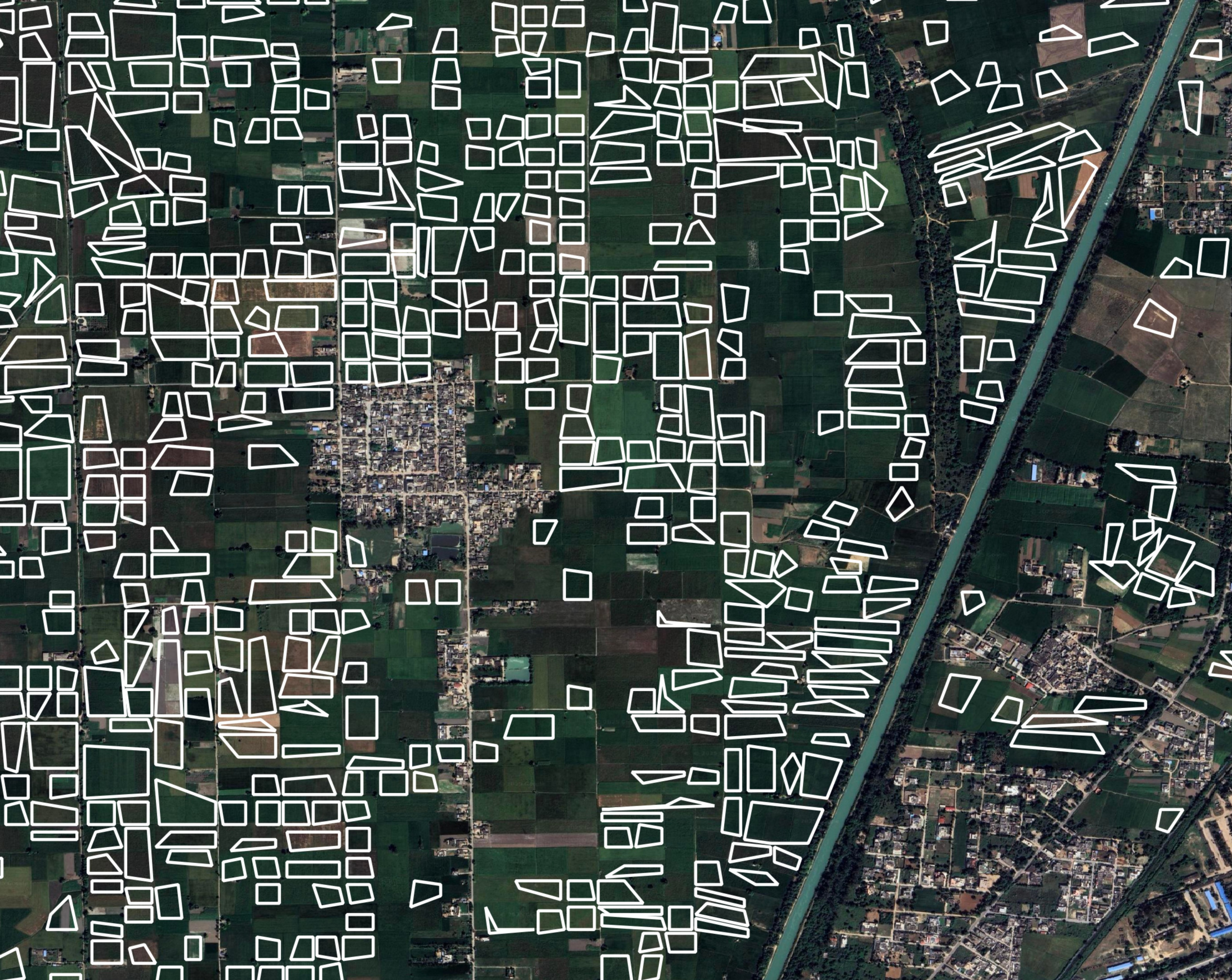}
        \caption{Ludhiana district. Rural area with a canal running through the eastern end.}
        \label{fig:sup:ftw_ludhiana}
    \end{subfigure}

    \vspace{0.3cm} 

    \begin{subfigure}[b]{\linewidth}
        \centering
        \includegraphics[width=\linewidth]{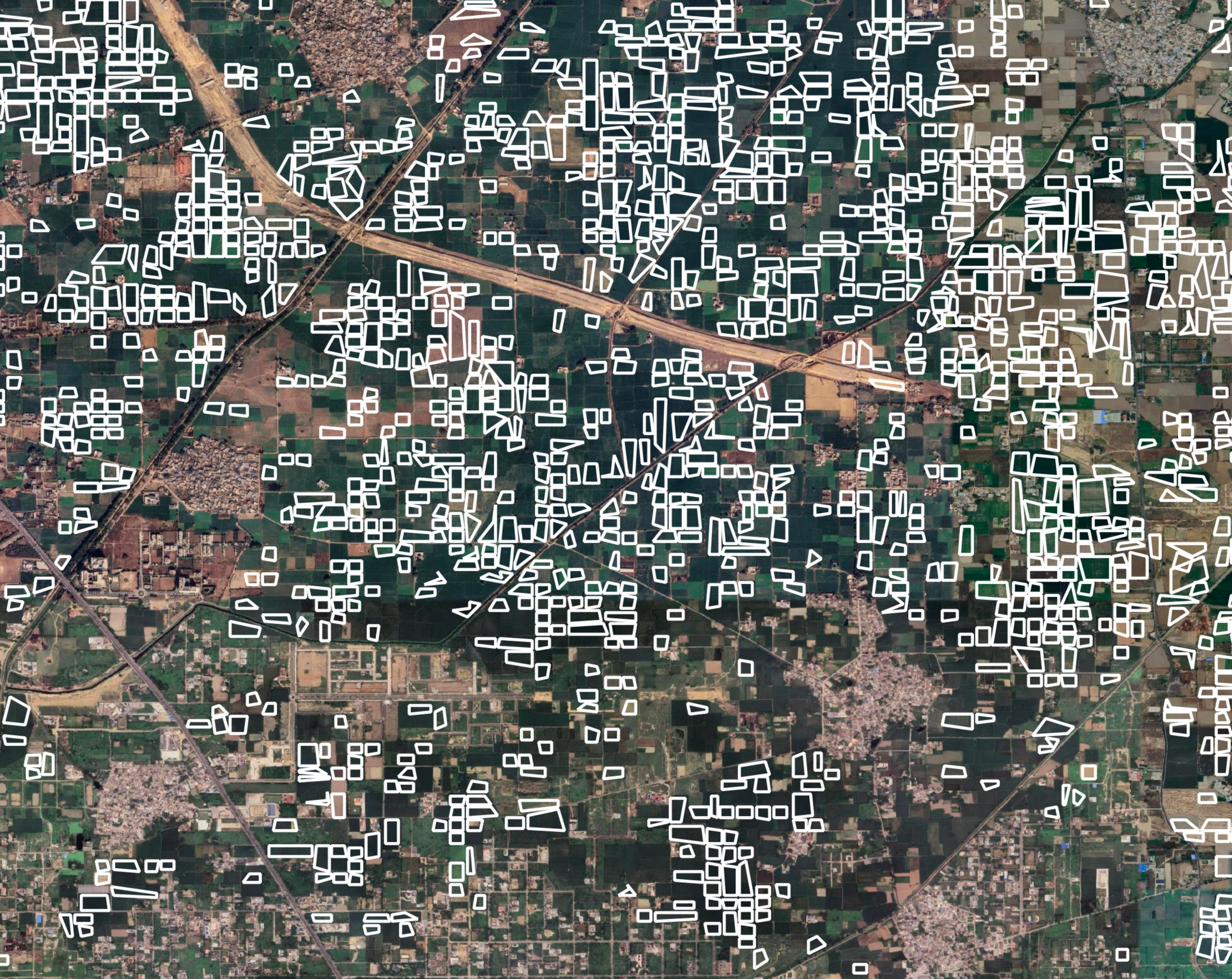}
        \caption{Amristar district. Semi-rural area outside the city of Amritsar}
        \label{fig:sup:ftw_fazilka}
    \end{subfigure}

    \caption{FTW delineation performance samples. Polygons in white represent FTW polygons, basemap from Google Maps Satellite view. Note that FTW performs relatively well in these samples and generally for plots of this size. Overall it tends to underestimate the number of plots and the size of the plot as well as have boundary artifacts.}
    \label{fig:sup:ftw_insets}
\end{figure}

Thereafter, we follow the method laid out in Fig.~\ref{fig:sup:workflow} with ascending orbit only, and use the best pretrained model for DSR classification, without any climatology inputs. The combination of pretrained models such as Presto and FTW, along with our dimensional training methodology that requires no cropping calendar information, allows us to apply this methodology to arbitrary spatial scales. For samples of the output produced over Punjab, refer to \Cref{fig:sup:ftw_insets}.

\begin{figure}[!t]
   \centering
   \includegraphics[width=\linewidth]{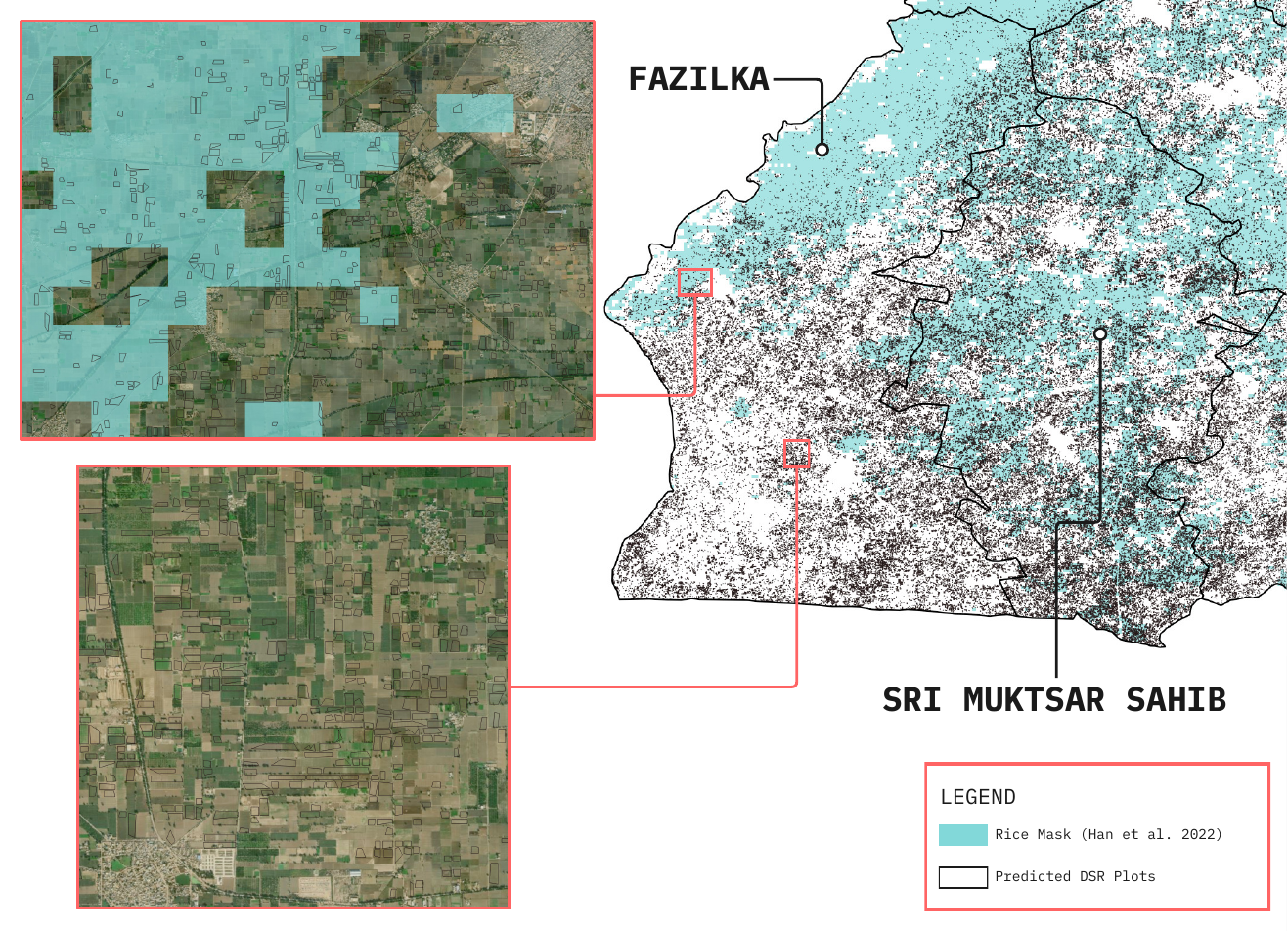}
   \caption{Outlier districts for masking: The two districts shown here have very low rice plot coverage according to \cite{han2022annual} relative to DSR mapping. The insets show both the coarse nature of the rice mask, generated at 500m resolution, as well as the potential undercounting of large areas of rice.}
   \label{fig:sup:outliers}
\end{figure}

\subsection{Validation}\label{sec:sup:val}

The task most relevant to policymaking for the PRANA program is to predict across the entire state to enable decision making at the district and sub-district level to be able to deploy resources towards increasing the adoption of DSR practices. However, the task of interpolating across roughly 3 million plots based on a model trained on just over 1,400 data points requires thorough validation. Manual validation of model estimates was beyond the scope of this project, and to produce some confidence at the district level, we aggregate the results and compare to the Government of Punjab's DSR incentive scheme. This involves providing a lump sum of Rs. 1,500 per acre to farmers who self-report acreage under DSR. For 2024, the total reported area under DSR cultivation was 253,328 acres~\cite{progressivefarming2025}. We compare these district-level estimates to~\citeauthor{han2022annual} as well as our estimates in~\Cref{tab:sup:comp-govt}. The column ``DSR Area'' is the raw sum of all DSR plots extracted per district. This aggregate area is almost certainly incorrect due to pixel-level errors in plot extraction from Sentinel-2 imagery. Furthermore, we did not implement a plot-level filter for rice v/s non-rice growing plots, since our training dataset did not consist of any non-rice plots in the kharif season. Therefore, we apply the rice growing regions as a mask from Han et. al 2022, and the estimates of DSR acreage are shown under the column ``DSR Area (masked)''.

\begin{figure*}[!htbp]
   \centering
   \includegraphics[width=\textwidth]{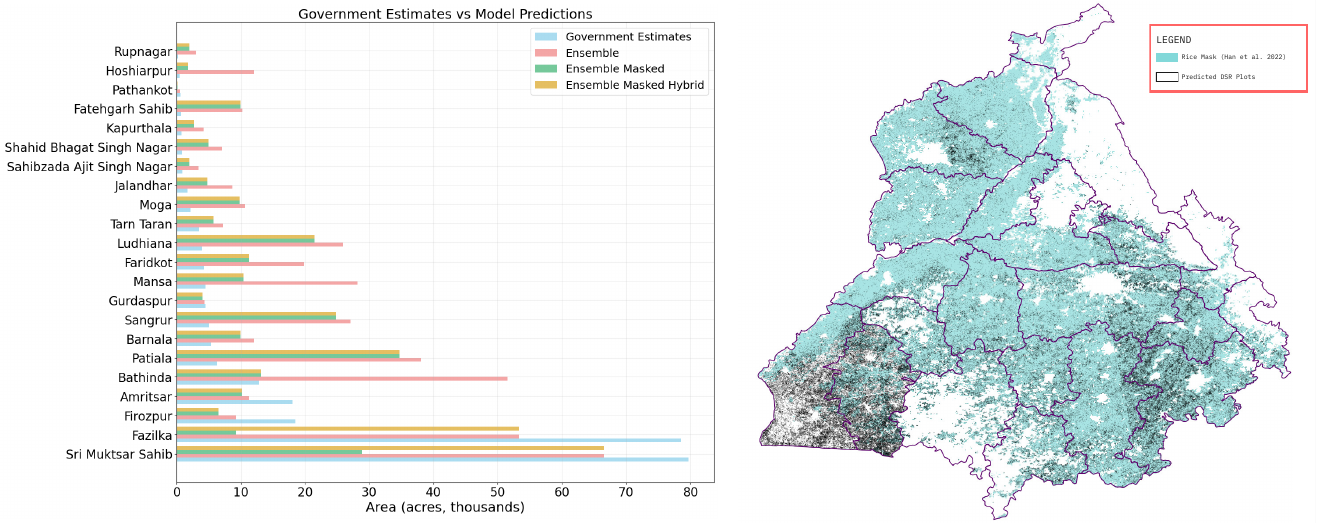}
   \caption{Comparison of different masking methodologies used in this study. Left: `Ensemble' refers to the DSR ensemble model outputs. `Ensemble Masked' is the ensemble masked with \cite{han2022annual}. `Ensemble Masked Hybrid' is where we use the outputs from Ensemble Masked except for Fazilka and Sri Muktsar Sahib. Right: The final DSR plot distributions from Ensemble Masked Hybrid overlaid on Han et al. 2022's estimates for rice producing regions in the year 2021.}
   \label{fig:sup:validation-comp}
\end{figure*}

The overestimation relative to government estimates reduces in the masked column, but we see a extremely large change in our DSR estimates for the districts of Sri Muktsar Sahib and Fazilka (highlighted rows in~\Cref{tab:sup:comp-govt}), with changes in acreage in the order of 40k. Further analysis (see~\Cref{fig:sup:outliers}) revealed that rice growing regional estimates~\cite{han2022annual} made for 2021 are likely underestimating outcomes for 2024, given that the state of Punjab has been a rapid increase in rice cultivation areas. This the south-west region which has seen more than a doubling of rice-growing regions in the last two decades~\cite{kaur2025rice}. We therefore created a hybrid version of our estimates and the masked estimates by keeping all districts masked except for Fazilka and Sri Muktsar Sahib. The comparisons of district-wise acreage across these three methods is shown in \Cref{fig:sup:validation-comp} (left). On the right of \Cref{fig:sup:validation-comp} we show hybrid DSR plot coverage along with the rice growing regions from~\cite{han2022annual}. These two districts are also the driest regions and therefore the need to implement water-saving methods would be the highest.

\begin{figure*}[tbhp]
    \centering
    \begin{subfigure}[b]{0.49\textwidth}
        \centering
        \includegraphics[width=\linewidth]{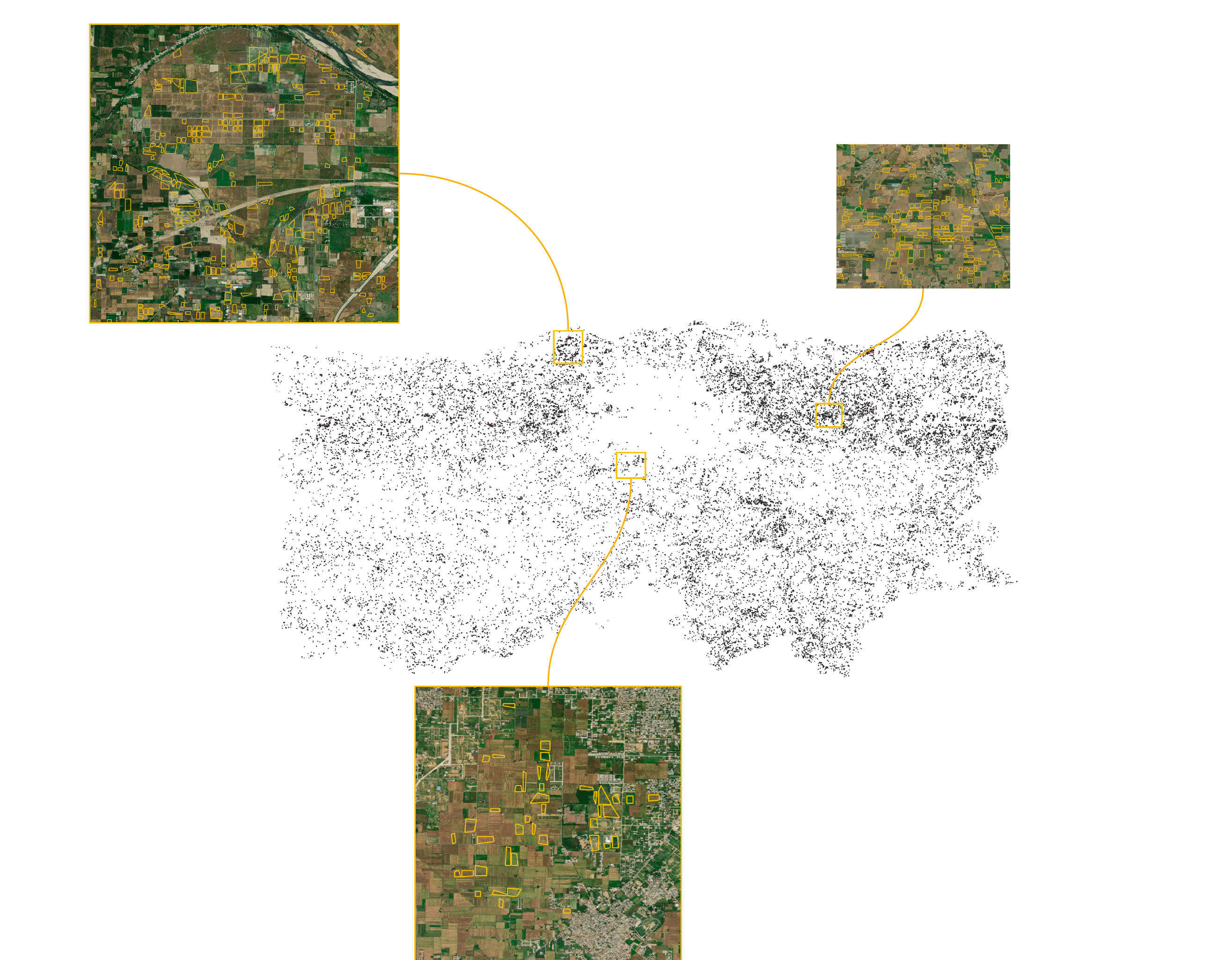}
        \caption{Ludhiana district}
    \end{subfigure}
    \hfill
    \begin{subfigure}[b]{0.49\textwidth}
        \centering
        \includegraphics[width=\linewidth]{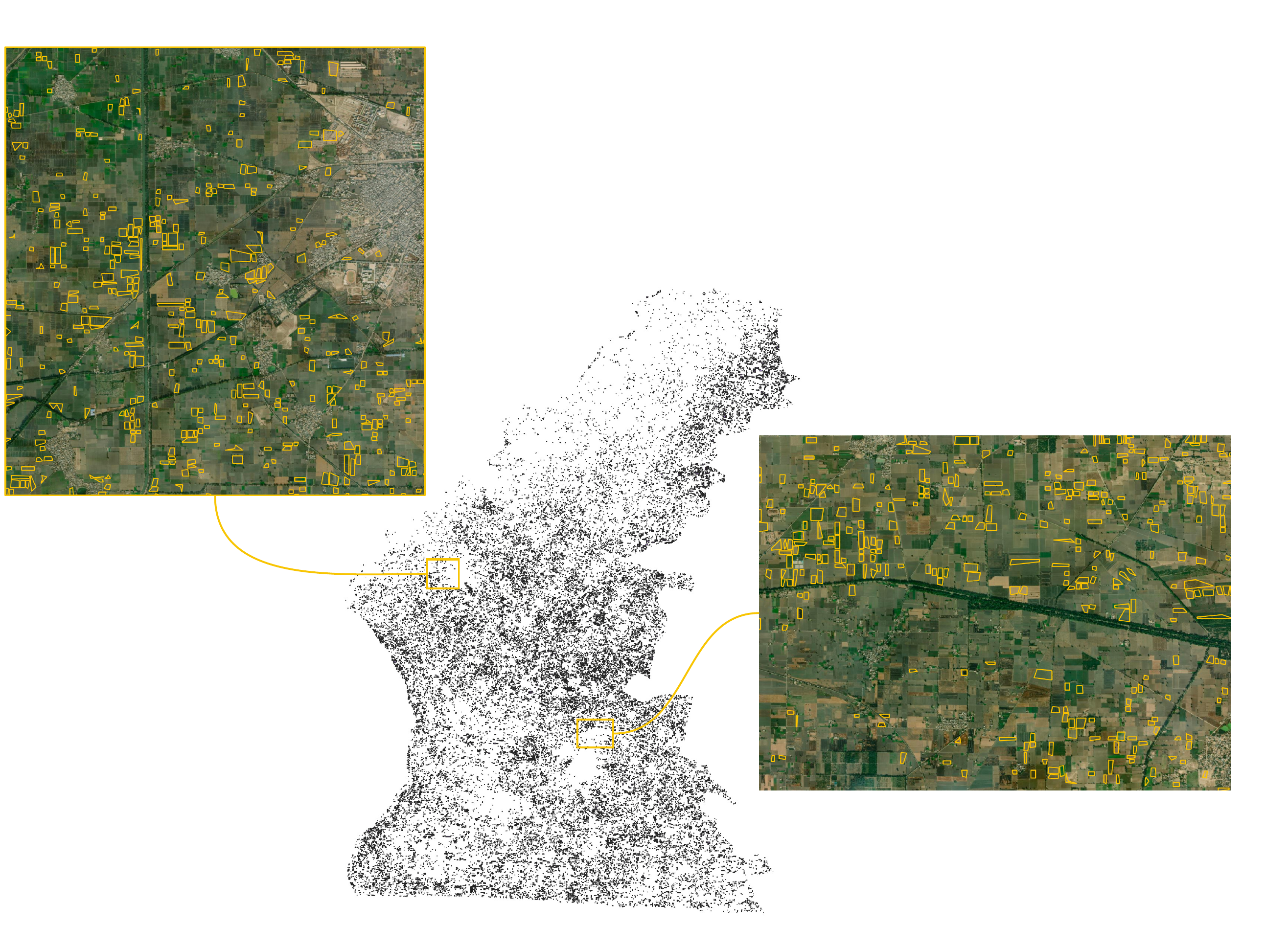}
        \caption{Fazilka district}
    \end{subfigure}

    \vspace{0.3cm}

    \begin{subfigure}[b]{0.49\textwidth}
        \centering
        \includegraphics[width=\linewidth]{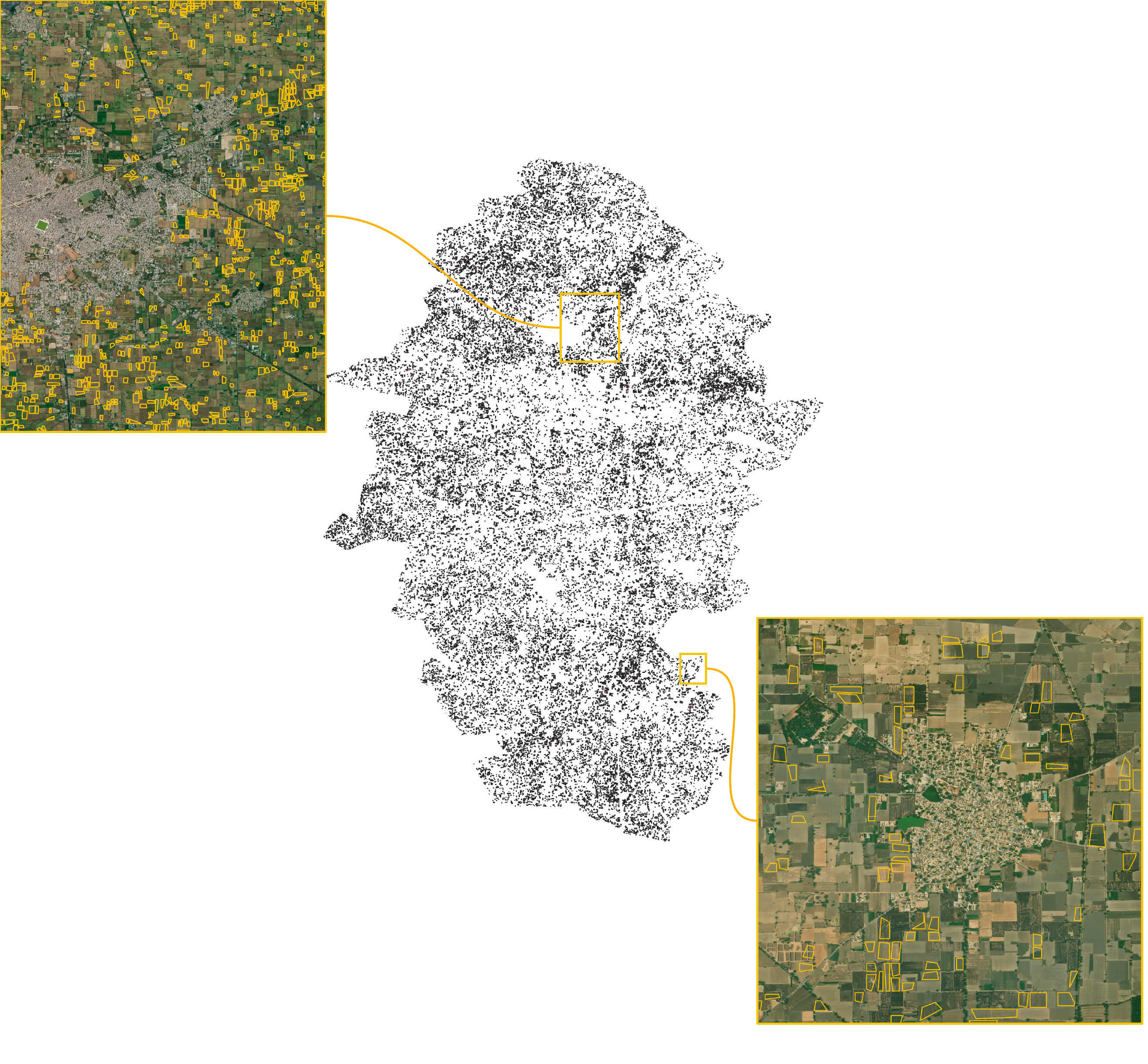}
        \caption{Sri Muktsar Sahib district}
    \end{subfigure}
    \hfill
    \begin{subfigure}[b]{0.49\textwidth}
        \centering
        \includegraphics[width=\linewidth]{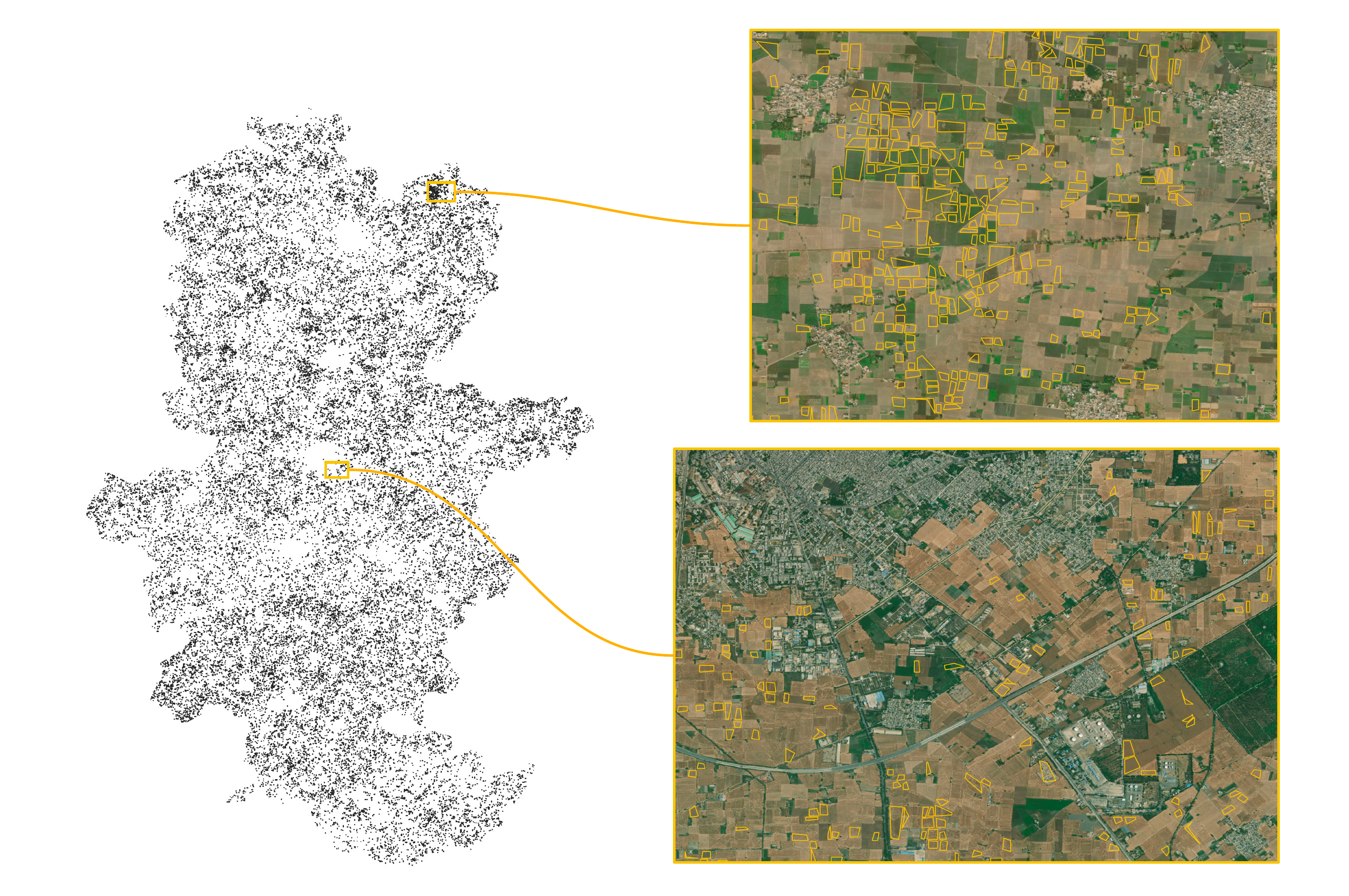}
        \caption{Sangrur district}
    \end{subfigure}

    \caption{DSR Plot predictions for different districts in Punjab, with insets showing sample crops of localized regions}
    \label{fig:dsr-preds}
\end{figure*}

For qualitative analysis of DSR plot predictions, please refer to \Cref{fig:dsr-preds}.

\begin{table*}[!htbp]
\centering
\resizebox{\textwidth}{!}{%
\begin{tabular}{@{}lccccccc@{}}
\toprule
Estimator (acres $\rightarrow$) &
  ~\cite{han2022annual} &
  \multicolumn{2}{c}{Government Estimates} &
  \multicolumn{4}{c}{Our estimates} \\
  \cmidrule(lr){1-1} \cmidrule(lr){2-2} \cmidrule(lr){3-4} \cmidrule(lr){5-8}
District ($\downarrow$) &
  \multicolumn{1}{l}{Rice Growing Area (2021)} &
  \multicolumn{1}{l}{DSR Estimates} &
  \multicolumn{1}{l}{\% under DSR} &
  \multicolumn{1}{l}{DSR Area} &
  \multicolumn{1}{l}{\% under DSR} &
  \multicolumn{1}{l}{DSR Area (masked)} &
  \multicolumn{1}{l}{\% under DSR} \\
  \cmidrule(lr){1-1} \cmidrule(lr){2-2} \cmidrule(lr){3-4} \cmidrule(lr){5-8}
Amritsar                   & 524706 & 18049 & 3.4\%  & 11265 & 2.1\%  & 10150 & 1.9\% \\
Barnala                    & 304892 & 5380  & 1.8\%  & 12036 & 3.9\%  & 9942  & 3.3\% \\
Bathinda                   & 283112 & 12786 & 4.5\%  & 51532 & 18.2\% & 13138 & 4.6\% \\
Faridkot                   & 229485 & 4330  & 1.9\%  & 19852 & 8.7\%  & 11283 & 4.9\% \\
Fatehgarh Sahib            & 267082 & 663   & 0.2\%  & 10251 & 3.8\%  & 9970  & 3.7\% \\
\rowcolor[HTML]{FFE599} 
Fazilka                    & 259095 & 78558 & 30.3\% & 53303 & 20.6\% & 9304  & 3.6\% \\
Firozpur                   & 410187 & 18452 & 4.5\%  & 9287  & 2.3\%  & 6554  & 1.6\% \\
Gurdaspur                  & 538681 & 4527  & 0.8\%  & 4349  & 0.8\%  & 4013  & 0.7\% \\
Hoshiarpur                 & 132811 & 501   & 0.4\%  & 12073 & 9.1\%  & 1774  & 1.3\% \\
Jalandhar                  & 398008 & 1673  & 0.4\%  & 8683  & 2.2\%  & 4764  & 1.2\% \\
Kapurthala                 & 306852 & 771   & 0.3\%  & 4249  & 1.4\%  & 2717  & 0.9\% \\
Ludhiana                   & 727297 & 3928  & 0.5\%  & 25898 & 3.6\%  & 21506 & 3.0\% \\
Mansa                      & 260315 & 4495  & 1.7\%  & 28184 & 10.8\% & 10407 & 4.0\% \\
Moga                       & 522039 & 2183  & 0.4\%  & 10660 & 2.0\%  & 9824  & 1.9\% \\
\rowcolor[HTML]{FFE599} 
Sri Muktsar Sahib          & 317697 & 79720 & 25.1\% & 66546 & 20.9\% & 28833 & 9.1\% \\
Pathankot                  & 56211  & 628   & 1.1\%  & 545   & 1.0\%  & 105   & 0.2\% \\
Patiala                    & 679038 & 6267  & 0.9\%  & 38061 & 5.6\%  & 34677 & 5.1\% \\
Rupnagar                   & 103007 & 232   & 0.2\%  & 3020  & 2.9\%  & 2002  & 1.9\% \\
Sahibzada Ajit Singh Nagar & 96121  & 914   & 1.0\%  & 3395  & 3.5\%  & 1992  & 2.1\% \\
Sangrur                    & 795607 & 5017  & 0.6\%  & 27097 & 3.4\%  & 24803 & 3.1\% \\
Shahid Bhagat Singh Nagar  & 170241 & 808   & 0.5\%  & 7082  & 4.2\%  & 4967  & 2.9\% \\
Tarn Taran                 & 531500 & 3446  & 0.6\%  & 7236  & 1.4\%  & 5758  & 1.1\% \\ \bottomrule
\end{tabular}%
}
\caption{Comparison of different rice growing area estimates. All areas are in acres.}
\label{tab:sup:comp-govt}
\end{table*}

\end{document}